%% file: main.tex
\documentclass[final]{cvpr}

\usepackage{times}
\usepackage{epsfig}
\usepackage{graphicx}
\usepackage{amsmath}
\usepackage{amssymb}

\usepackage{xcolor}
\usepackage{subcaption}
\usepackage{overpic}
\usepackage{wrapfig}
\usepackage{soul}
\usepackage[toc,page]{appendix}
\usepackage{xr}
\usepackage[pagebackref=true,breaklinks=true,colorlinks,bookmarks=false]{hyperref}

\begin{document}

\title{3D Shape Generation with Grid-based Implicit Functions}

\author{%
{\parbox{\textwidth}{\centering \hspace{\stretch{2}} Moritz Ibing\hspace{\stretch{1.5}}Isaak Lim\hspace{\stretch{1.5}}Leif Kobbelt\hspace{\stretch{2}} }} \\
  Visual Computing Institute, RWTH Aachen University
}

\maketitle

\begin{abstract}
    Previous approaches to generate shapes in a 3D setting train a GAN on the latent space of an autoencoder (AE). Even though this produces convincing results, it has two major shortcomings.
    As the GAN is limited to reproduce the dataset the AE was trained on, we cannot reuse a trained AE for novel data. Furthermore, it is difficult to add spatial supervision into the generation process, as the AE only gives us a global representation.
    To remedy these issues, we propose to train the GAN on grids (i.e.\ each cell covers a part of a shape).
    In this representation each cell is equipped with a latent vector provided by an AE. 
    This localized representation enables more expressiveness (since the cell-based latent vectors can be combined in novel ways) as well as spatial control of the generation process (e.g.\ via bounding boxes).
    Our method outperforms the current state of the art on all established evaluation measures, proposed for quantitatively evaluating the generative capabilities of GANs. We show limitations of these measures and propose the adaptation of a robust criterion from statistical analysis as an alternative. 
\end{abstract}

\section{Introduction}
Training stable 3D GANs can be challenging and often better results are obtained by splitting the generation process into two parts. First an autoencoder (AE) is trained to obtain a compressed latent representation, then a GAN is trained to model the density of this global latent space. This simplifies the task, as the AE typically generates reasonable output for a wide range of latent vectors.
Training the AE can be seen as imposing a bias on the generation process to produce shapes similar to those seen during its training.

This procedure has obtained convincing results both for the generation of point clouds \cite{Achlioptas2018LearningRA} and implicit functions \cite{chen2019implicit_decoder}. However, using the latent space of an AE as intermediate shape representation has some severe drawbacks. Small variations of shapes are usually of local nature. As the latent representation used in prior work is of inherently global nature, such changes cannot be modeled easily. This makes it difficult to add any localized modifications or constraints into the generation process. Tasks such as conditional generation based on semantic information (e.g.\ segmentation/part labels) become hard to train. 
Another problem is that the AE limits the space of shapes that can be generated. Indeed every AE, which does not represent the entire space of possible shapes in its latent space, limits the expressiveness of the GAN through the bias it imposes. E.g.\ if we train a GAN to create tables in such a manner, this is unlikely to work with the latent space of an AE trained for chairs.

A more natural choice of representation is to view a shape as a composition of many different local parts. This allows the network to choose parts that occur in different objects and arrange them to create new shapes.
Recently, Jiang et al.~\cite{jiang2020local} learned local latent representations by subdividing the object space into a grid to reconstruct scenes from point clouds with high fidelity.
This can be seen as only imposing a localized bias on the shape generation process.
Learning to capture the geometry within a single grid cell is much simpler than learning to represent the whole shape. Thus the reconstruction quality of the AE is much improved. 

We argue that this representation has several further advantages.
Training a generative model on this  localized latent space enables more variety in the generated shapes, as latent vectors belonging to different shapes or even different object classes can be mixed, giving the generator more degrees of freedom. Its expressiveness is thus not limited by the AEs ability to generalize. In fact we do not even need to train an AE for each specific class. An AE trained on cells of e.g.\ tables can still be used to generate chairs, as on part level both classes share similarities.
Furthermore, we can use convolutional architectures and thus build on existing research in image generation. Although this argument applies to voxel based models as well~\cite{wu2016learning}, those are limited by the grid resolution that can fit into memory. As we can represent complex surfaces per grid cell, we can model realistic shapes while still keeping the grid resolution low.
Lastly, the spatial decomposition of the latent space allows us to perform conditional generation based on spatial information, such as bounding boxes or semantic labels.

When it comes to evaluating GANs there is no universally established measure to quantitatively rate, to what degree such networks are able to approximate a data distribution.
Image GANs are usually evaluated with the inception score~\cite{salimans2016improved} or Fréchet inception distance~\cite{heusel2017gans}. The computation of both measures involves the application of the inception network~\cite{szegedy2016rethinking}. Since many different representations are used to encode 3D shapes (point clouds, voxel grids, implicit functions, etc.), this is not straightforward to apply for our use case.
While previous quality scores for 3D data have been proposed, we show that they have some limitations. Therefore, we propose to apply a statistic measure from a two sample test for multivariate sets~\cite{chen2017new} to compute the statistical difference between a set of generated objects and a test set of unseen shapes from the data distribution.

Our key contributions are as follows:
\begin{itemize}
    \item We show that localized grid-based implicit functions are better suited for 3D shape generation with GANs than global implicit functions for three reasons. First, they offer higher quality results, since each cell only has to represent fairly simple geometry. Second, localized implicit functions can be combined together in novel ways and therefore offer more flexibility. Third, localized grid-based implicit functions allow us to control the generation process spatially, which is difficult to accomplish with global implicit functions.
    \item We show that common evaluation techniques for 3D shape generation have several drawbacks. To alleviate this we propose a new robust score inspired from statistical analysis.
\end{itemize}

\section{Related Work}
Many different representations (and network architectures) for processing 3D shapes have been proposed. In the context of this paper we will focus on previous work on the representation of shapes as functions via neural networks as well as prior work on generation of shapes with GANs.

\subsection{Representing Shapes as Functions}
There are two families of functions that can be used to represent shapes: parametric and implicit functions.
The parametric approach to describe a surface (preferably an orientable 2-manifold in 3D) is to define a function $\mathbb{R}^2 \rightarrow \mathbb{R}^3$. Evaluating this function (e.g.\ on a 2D unit square) then gives positions on the surface of a shape in $\mathbb{R}^3$. Such maps can be learned and represented by neural networks~\cite{sinha2017surfnet}. Since a single map is often insufficient to represent complex shapes, this idea has been modified in AtlasNet \cite{atlas18} to instead learn several functions, where each describes part of a single shape. A different approach is taken by~\cite{lim2019convolutional} who instead represent a shape with localized functions in a grid, where each function models the surface within a grid cell.

Implicit functions come in two types: signed distance functions $\mathbb{R}^3 \rightarrow \mathbb{R}$ map each point in 3D to a distance value, that is by convention negative inside of the shape and positive outside. The surface of the shape is then the zero level-set of this function. On the other hand there are binary functions $\mathbb{R}^3 \rightarrow \{0,1\}$, that classify a point as being inside or outside of a given shape. Methods such as Marching Cubes \cite{lorensen1987marching} can be used in both cases to extract a surface. Recently several approaches have been published, that apply neural networks to represent such functions~\cite{park2019deepsdf, mescheder2019occ, chen2019implicit_decoder}. As a global function can have difficulties representing a shape with all its details, several methods have been proposed to mitigate this problem by using localized functions embedded in a grid structure. Chibane et al.~\cite{chibane2020implicit} employ a multi-scale approach, where feature vectors at different scales and grid points are interpolated based on the current point position, while Jiang et al.~\cite{jiang2020local} save a single feature vector per grid cell and classify a point (in local coordinates) together with its cell vector. We use a similar approach for our generative model.

\subsection{GANs in 3D}
To the best of our knowledge \cite{wu2016learning} were the first to develop GANs for the 3D setting, choosing voxels as a representation. This allowed them to use convolutional networks both as the generator and discriminator. However, the results generated by their method are limited by the low resolution of $64^3$ grids (mainly due to memory constraints).
Achlioptas et al.~\cite{Achlioptas2018LearningRA} developed GANs for point clouds in two settings. They trained a generator and discriminator directly on the point clouds but also introduced a two step approach. They first trained an autoencoder for reconstructing point clouds and then fitted a generative model to the latent space. For this they proposed both a GAN and a Gaussian Mixture Model.
A similar approach is used by \cite{LiZZPS19}. 
The idea of splitting GAN training into two steps proved to be applicable to implicit functions as well, as IM-GAN \cite{chen2019implicit_decoder} demonstrated. Very recently \cite{kleineberg2020adversarial} showed that it is also possible to train GANs directly on implicit functions, however they do not reach the same quality as latent approaches.

\section{3D Shape Generation with Localized Implicit Functions}

\begin{figure*}
  \centering
  \includegraphics[width=0.9\linewidth]{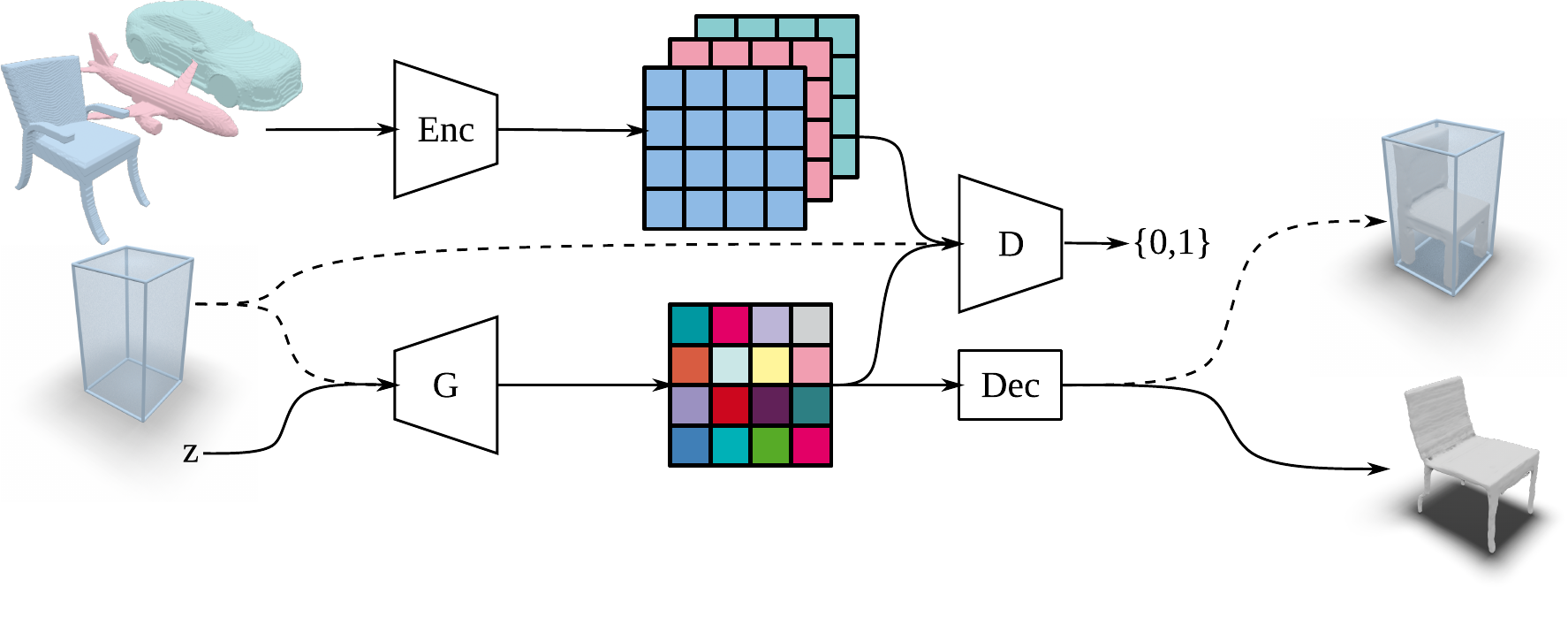}
      \caption{Overview of our method: With the help of an AE (Enc \& Dec) we can represent 3D shapes as a grid of latent vectors. Our proposed GAN (G \& D) generates new grids of latent vectors, from which shapes can be extracted via the decoder (Dec). Furthermore, we can optionally (dashed arrows) condition, e.g. on bounding boxes, for spatial control in the generation process.}
  \label{fig:overview}
\end{figure*}

We propose a two step procedure for the generation of 3D shapes (see Figure~\ref{fig:overview}) similar to \cite{chen2019implicit_decoder}. Firstly, we train an autoencoder that maps inputs to localized implicit functions. Secondly, we train a GAN that learns the distribution of such functions. We will discuss our autoencoder first and then describe our proposed unconditional and conditional generation methods. Details of the different architectures and the training procedure can be found in the supplementary material (Appendix \ref{sec:arch}, \ref{sec:training}).

\subsection{Autoencoder}
The architecture we employ for our autoencoder is related to~\cite{jiang2020local}.
Our encoder takes as input high-resolution binary voxel grids ($n^3$) and after several (strided) convolutions outputs (comparably) low-resolution grids ($k^3$), where each grid cell is equipped with an $h$-dimensional latent vector. Thus each cell of the output encodes the geometry of an overlapping cubical subset of the high-resolution voxel grid. In all of our experiments the resolution of the latent grid is chosen as $k$ = 32 and the number of channels as $h$ = 8. 

The decoder evaluates the implicit function represented by this grid of latent cells at a set of sample points. To this end, we find the cell a point is contained in and concatenate its latent vector with the point's position (in local coordinates of the cell). Then we feed this vector into a MLP with a scalar output. The resulting score (0 for outside and 1 for inside of a shape) can be trained with a binary cross entropy loss.

As independent functions in neighboring cells can lead to discontinuities of the implicit function at the boundary between cells, we interpolate the results obtained from neighboring grid cells, leading to smoother results.
\begin{equation}\label{eq:smoothing}
  f(x, C, \theta) = \sum_{j \in \mathcal{N}} w_j \operatorname{Dec}_\theta(c_j \in C, k(x - x_j)),
\end{equation}
where $C$ is the grid of latent vectors, $c_j$ is the vector corresponding to cell $j$ and $x_j$ is its cell center, $\mathcal{N}$ is the neighborhood of point $x$ and $w_j$ the trilinear interpolation weight of $x$ with regard to the center of cell $j$. We refer to the resolution of the latent grid as $k$. The decoder $\operatorname{Dec}$ has parameters $\theta$.
This interpolation is especially important in our GAN setting, as it simplifies the task of generating matching cells.
To make sure that the necessary information for this is stored in the latent vectors, we choose the receptive field of the encoder so that each latent grid cell can encode the geometric information of neighboring cells as well.

\subsection{Unconditional Generation}
After we have trained an autoencoder, as described in the previous section, we can now train a GAN on its latent space. Chen et al.~\cite{chen2019implicit_decoder} learn a single latent vector for the entire shape and therefore train simple MLPs as generator and discriminator to generate new vectors and thus new shapes.
We on the other hand generate a full grid of latent vectors, each describing a local part of the shape, that together constitute the object. Thus, it is not only important that the local geometry described by the implicit function for each grid cell itself is reasonable, it must match with its neighbors as well.
Although these considerations put an additional burden on the GAN, this approach has a decisive advantage over previous latent approaches. As it is possible to arrange the grid cells in various ways, we are less constrained by any bias the autoencoder may introduce. This is because we only need the autoencoder to reconstruct building blocks instead of entire shapes.

Our task can be seen as being analogous to image generation, therefore similar considerations apply. Instead of generating a 2D grid with 3 channels, we generate a 3D grid with $h$ channels. 
For this reason we use 3D convolutional networks as generator and discriminator. 
The generator is a simple 3D CNN with strided convolutions, batch normalization and LeakyReLU as non-linearity. We do not use any residual layers or skip connections. 

In order to provide feedback to the generator at various scales, we use three patch discriminators (adapted from pix2pix~\cite{isola2017image}) that have residual layers and apply spectral normalization \cite{miyato2018spectral}. 
All three discriminators have the same architecture, but do not share parameters. Each discriminator is run convolutionally over the latent grid, rating each local patch separately. The scores are then averaged for the discriminators' final output. The discriminators work at different input resolutions ($32^3$, $16^3$, and $8^3$) and thus inspect features of different scale. The lower resolutions are computed by trilinear interpolation of the latent vectors. 
This training scheme puts a strong focus on local details, emphasizing the higher complexity and expressiveness of single grid cells compared to images or binary voxel grids. The effectiveness of patch discriminators is shown in Appendix \ref{sec:evaluation}.

For better convergence we use a gradient penalty on interpolations between true and fake data as introduced in \cite{gulrajani2017improved}. We obtained the best results with zero-centered gradient penalty \cite{roth2017stabilizing, mescheder2018training} and maximum reduction \cite{jolicoeur2019connections}. As loss we use the standard non-saturating (NS) loss \cite{goodfellow2014generative}.

\subsection{Conditional Generation}
\label{sec:conditional}
In many use cases we would like to control certain characteristics of the generated objects. To this end we can guide the generation process with conditional GANs \cite{mirza2014conditional}.
These conditions can be of global nature, like class labels, or more localized. An example for this are architectures that  create images based on information in the form of other images (e.g.\ photos from label maps) \cite{isola2017image, wang2018high, park2019semantic}. For this purpose the input image is used as a mask providing a pixel-wise conditioning.

The spatial organisation of our latent space easily allows us to adapt this approach to 3D data. For our generator we use the architecture of SPADE \cite{park2019semantic} adapted to 3D. The key idea is to compute a cell-wise scale and bias depending on a mask and apply it to the feature maps at different layers.
We use the same discriminators as for unconditional generation. The only difference is that we concatenate the mask with the generated grid in the beginning. We experiment with different ways to provide these masks (e.g. bounding boxes, shape parts, silhouettes), depending on the respective application. The losses and training parameters are the same as for unconditional generation.

\section{Evaluation}
In this section we present the measure we use to compare the performance of different GANs as well as motivate its effectiveness compared to previous measures. Furthermore, we show quantitative and qualitative results of our unconditional and conditional generation scheme.
All evaluations are performed on the ShapeNet Core dataset (v1) \cite{chang2015shapenet}. For comparability reasons we follow the training split and evaluation setup from~\cite{chen2019implicit_decoder} and generate distributions for the categories car, chair, plane, rifle and table. As ground truth we use the voxelized models from \cite{hane2017hierarchical}. In each category we sorted the models by name and used the first 80\% for training and the rest for testing.

\subsection{Quality Measures}

\begin{figure*}
  \centering
  \includegraphics[width=0.32\linewidth]{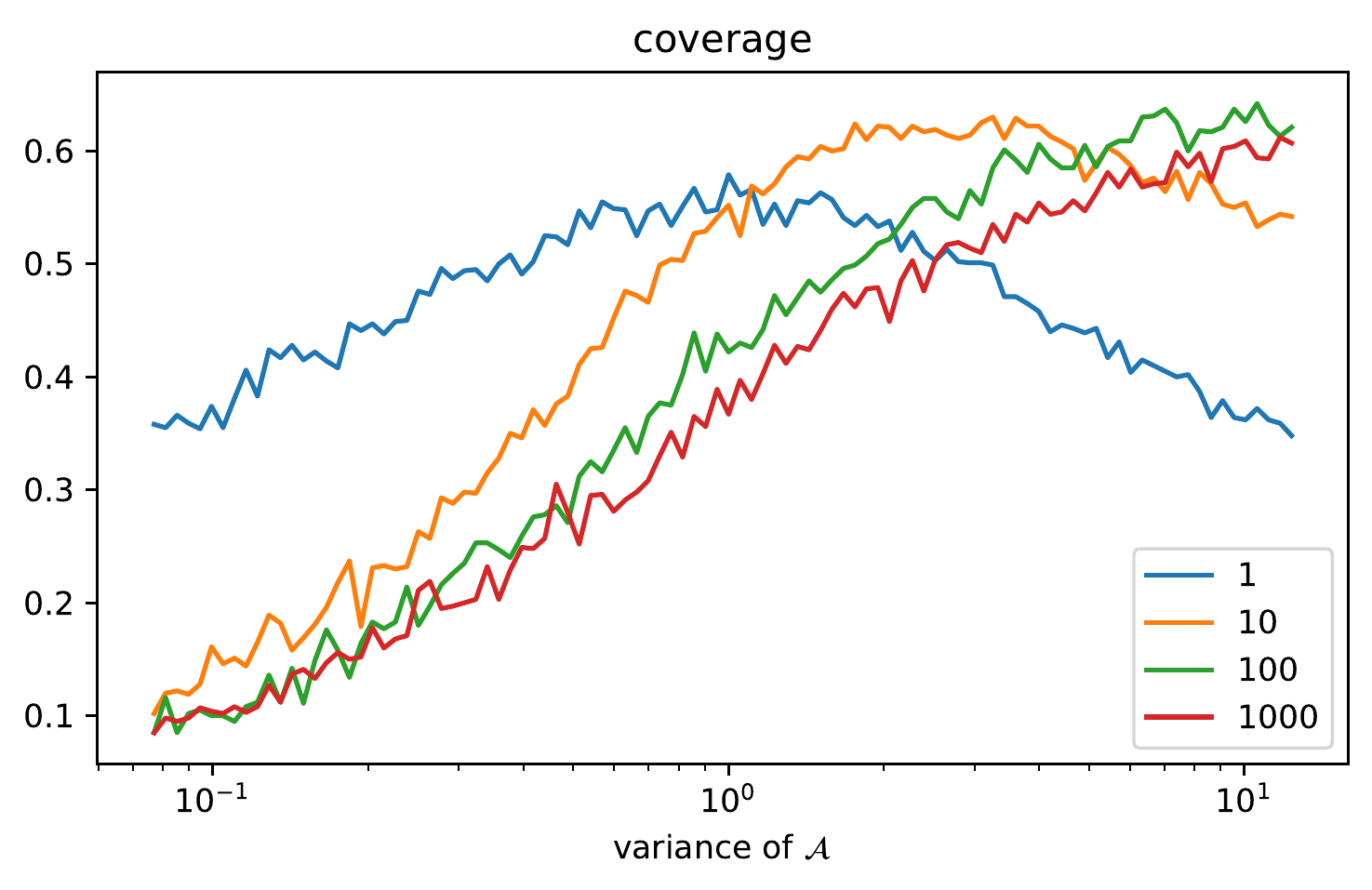}
  \includegraphics[width=0.32\linewidth]{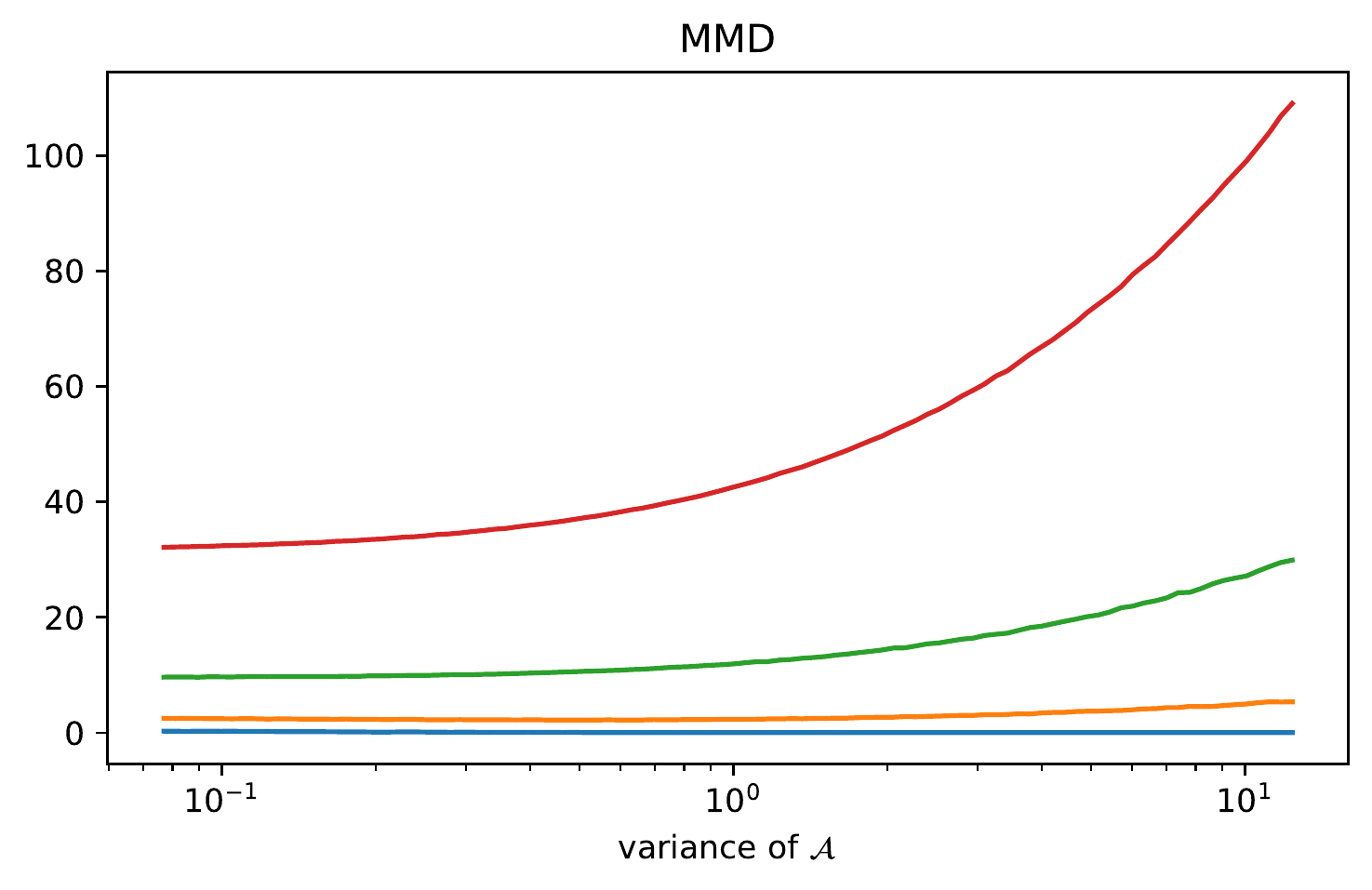}
  \includegraphics[width=0.32\linewidth]{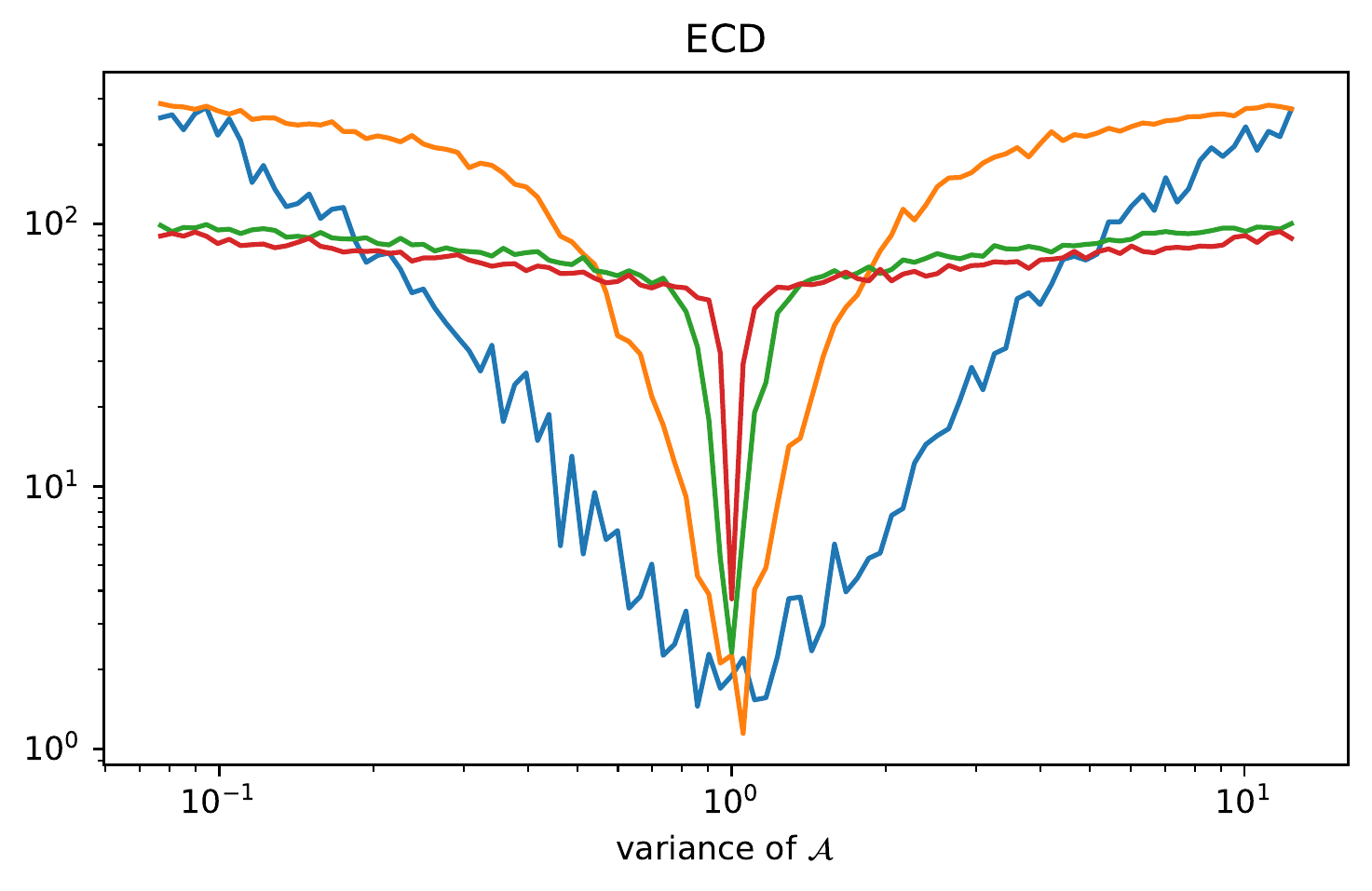}
  \caption{Comparing different distribution measures on two Gaussian distributions with different dimensions (1, 10, 100, 1000) and zero mean. $\mathcal{B}$ has a variance of 1, while the variance of $\mathcal{A}$ varies.}
  \label{fig:measures}
\end{figure*}

Several approaches have been suggested to evaluate the fidelity of generative models. Generally any evaluation method needs to answer two questions:
\begin{enumerate}
    \item How to measure the similarity of individual data points (e.g. 3D shapes)?
    \item How to compare two data distributions?
\end{enumerate}
The answer to the first question typically depends on the domain, while the answer to the second question is domain agnostic for the most part.
In the realm of 3D shapes, several similarity measures have been proposed. For point cloud based methods the Chamfer distance (CD) and Earth Mover distance (EMD) were introduced \cite{Achlioptas2018LearningRA}. On the other hand \cite{shu20193d} proposes to use features extracted from a pretrained PointNet network \cite{qi2017pointnet}.
As we generate surfaces instead of point clouds, using these methods would require to sample all shapes, by which we lose fine details of the underlying surface. 
Chen et al.~\cite{chen2019implicit_decoder} argue that point based distances do not align well with visual similarity and instead propose to use the light field descriptor (LFD) \cite{chen2003visual} to measure similarity. We follow this reasoning and use the LFD for our evaluation. Note that the similarity between LFDs is measured in a non-Euclidean manner.

Having answered the first question, i.e.\ how to compare individual data points, we will now discuss how two distributions can be compared.
We use the test set as a proxy of a different unseen sampling drawn from the data distribution that generated the training set. 
We refer to the generated data set as $\mathcal{A}$ and to the test set as $\mathcal{B}$.

Two options for comparing distributions $\mathcal{A}$ and $\mathcal{B}$ were introduced by Achlioptas et al.~\cite{Achlioptas2018LearningRA}.
\emph{Coverage} is meant to measure the diversity of $\mathcal{A}$ (with regard to $\mathcal{B}$). For each shape in $\mathcal{A}$ we mark the closest neighbor in $\mathcal{B}$ according to the distance defined above. We then count the percentage of shapes in $\mathcal{B}$, that have been marked. If our generative scheme has low diversity, for example due to mode collapse, the generated shapes would lie close together and cover only a small subset of the space spanned by $\mathcal{B}$.
This measure however does not evaluate the fidelity of individual shapes within $\mathcal{A}$, as the actual distance between shapes does not matter. To take this into account the \emph{Minimum Matching Distance} (MMD) is introduced. Here we compute the distance from each shape in $\mathcal{B}$ to its closest neighbor in $\mathcal{A}$ and take the mean over those.

Another option to compare distributions that is popular for evaluating GANs on images, is the Fréchet inception distance (FID) \cite{heusel2017gans}. This measure assumes the distributions to be Gaussian and thus computes mean and covariance of the descriptors within $\mathcal{A}$ and $\mathcal{B}$ and then computes the Fréchet distance.
Note that this method assumes Euclidean distance between feature maps and does not work on general distance measures. Therefore, it cannot be applied to the LFD.

We argue that these metrics all have shortcomings in evaluating the quality of a generated distribution. For Coverage and MMD the outliers in $\mathcal{A}$ are not penalized in any way, whereas outliers in $\mathcal{B}$ can have a significant impact. Furthermore, considering only the nearest neighbor is not sufficient to compare the actual density of distributions, as differences in local densities do not have much of an impact. On the other hand FID runs into problems, when the underlying distributions are not Gaussian.

An easy example, where coverage and MMD already fail, is to compare (based on the Euclidean distance) samples from two isotropic Gaussian distributions with zero mean for various dimensions (Figure~\ref{fig:measures}). $\mathcal{B}$ was sampled from a Gaussian with variance 1, while we mimic different generative models by sampling multiple sets of $\mathcal{A}$ from different Gaussians with varying variance. Ideally, the measures would give the best results for $\mathcal{A}$ sampled from a Gaussian with variance 1.
However, the coverage never reaches 1 even when $\mathcal{A}$ and $\mathcal{B}$ are drawn from the same distribution. More critically,  depending on the dimension, the coverage does not have its peak at 1, as would be expected.
The MMD is either not much affected, or has its minimum, when the variance is minimal.
Thus, even in combination, the two measure are not reliable in order to determine how likely it is that the two sets come from the same distribution.

To show the deficiencies of the Fréchet distance as an evaluation measure, we compare different types of distributions, whose values we choose so that they all have zero mean and unit variance (Table~\ref{fid-bash}). By definition the Fréchet distance is not able to differentiate between any of these distributions, as the fitted Gaussians are identical.

\begin{table}
  \centering
  \begin{tabular}{|c|r|r|r|}
    \hline
                & Gaussian                  & uniform                   & binary          \\
    \hline
      Gaussian  & \textcolor{blue}{5.31}    & \textcolor{blue}{5.27}    & \textcolor{blue}{5.24}    \\
                & 1.55                      & 52.73                     & 163.58                    \\
      uniform   &                           & \textcolor{blue}{5.18}    & \textcolor{blue}{5.16}    \\
                &                           & 1.58                      & 135.21                    \\
      binary  &                           &                           & \textcolor{blue}{5.06}    \\
                &                           &                           & 1.95                      \\
    \hline
  \end{tabular}
  \caption{Comparing different measures on different distributions. Fréchet distance in blue, ECD in black. The samples are always 100 dimensional, where each value is chosen from a distribution with parameters chosen so that we have zero mean and unit variance. The binary distribution consists of the values -1 and 1.}
  \label{fid-bash}
\end{table}

\begin{table*}
  \begin{center}
    \begin{tabular}{|c|c|r|r|r|r|r|r|r|}
      \hline
                &           & Plane             & Car               & Chair             & Rifle             & Table             & Avg. w/o planes   & Avg.  \\
      \hline
        COV(\%) & 3DGAN     &                   & 12.13             & 25.07             & 62.32             & 18.80             & 29.58             &       \\
                & PC-GAN    & 73.55             & 61.40             & 70.06             & 61.47             & 77.50             & 67.61             & 68.80 \\
                & IM-GAN    & 70.33             & 69.33             & 75.44             & 65.26             & \textbf{86.43}    & 74.12             & 73.36 \\
                & Our       & \textbf{81.58}    & \textbf{80.67}    & \textbf{82.08}    & \textbf{81.47}    & 86.19             & \textbf{83.10}    & \textbf{82.80} \\
    \hline
                & Train     & 85.04             & 85.67             & 84.73             & 84.00             & 87.13              & 85.38            & 85.13 \\
    \hline
        MMD     & 3DGAN     &                   & 1,993             & 4,365             & 4,476             & 5,208             & 4,010             &       \\
                & PC-GAN    & 3,737             & 1,360             & 3,143             & 3,891             & 2,822             & 2,804             & 2,991 \\
                & IM-GAN    & 3,689             & 1,287             & 2,893             & 3,760             & 2,527             & 2,617             & 2,831 \\
                & Our       & \textbf{3,226}    & \textbf{1,225}    & \textbf{2,768}    & \textbf{3,366}    & \textbf{2,396}    & \textbf{2,453}    & \textbf{2,607}  \\
    \hline
                & Train     & 2,225             & 984               & 2,317             & 3,085             & 2,066             & 2,113             & 2,135 \\
      \hline
    \end{tabular}
  \end{center}
  \caption{Quantitative evaluation of generative models. As 3DGAN was not trained on plane models, this entry is missing. Results for the train set are reported to give reference values.}
  \label{results-table}
\end{table*}

\begin{table}
  \centering
  \begin{tabular}{|c|c|r|r|r|r|r|}
    \hline
            & Plane         & Car               & Chair         & Rifle         & Table         \\
    \hline
    3DGAN   &               & 28,855            & 26,279        & 6,495         & 32,116        \\
    IM-GAN  & 6,543         & 20,606            & 2,553         & 3,288         & 1,018         \\
    Our     & \textbf{355}  & \textbf{1,062}    & \textbf{144}  & \textbf{94}   & \textbf{188}  \\
    \hline
    Train   & 1             & 11                & 1             & 2             & 5             \\
    \hline
  \end{tabular}
  \caption{Quantitative evaluation of generative models with ECD. We do not report averages, as values for different dataset sizes are not comparable.}
  \label{ecd-table}
\end{table}

We therefore are interested in a test, that remedies these issues. It should take local densities into account, consider the complete distribution of $\mathcal{A}$ and $\mathcal{B}$ without being too much affected by outliers and actually decrease when the two distributions are similar. Furthermore it should be able to distinguish all kinds of distributions and not be restricted to Gaussians.
Testing whether two sample sets come from the same underlying distribution is a well known problem in statistical analysis and referred to as a \emph{two sample test}. 
Since we want to measure the distance between $\mathcal{A}$ and $\mathcal{B}$ by considering the likelihood of them being sampled from the same distribution, this problem is closely related to ours. 
We propose to use a statistic introduced in~\cite{chen2017new} as a distance measure between our two sets, since it is fairly robust even for multivariate data.
This approach builds a $k$-minimum spanning tree of the neighborhood graph of $\mathcal{A} \cup \mathcal{B}$. Edges in this tree are classified according to whether they connect shapes within $\mathcal{A}$, within $\mathcal{B}$ or between $\mathcal{A}$ and $\mathcal{B}$. The final score is computed as a weighted difference between the number of these edges and the edge count we would expect if $\mathcal{A}$ and $\mathcal{B}$ were from the same distribution. The exact formula for this measure can be found in Appendix \ref{sec:ecd}. Throughout the rest of this paper we will refer to this measure as \emph{Edge Count Difference} (ECD).

As shown in Figure \ref{fig:measures} this measure performs reliably on the toy example with Gaussian distributions introduced above. It actually achieves its minimum, when $\mathcal{A}$ and $\mathcal{B}$ are sampled from the same distribution. Furthermore it has no problems distinguishing different distributions with the same variance (Table \ref{fid-bash}). As we only need to be able to obtain distances between samples to compute the ECD, we are not restricted to the Euclidean space and thus can apply this test together with the LFD.

\subsection{Results}

\begin{figure*}
  \begin{subfigure}{\linewidth}
      \centering
      \hfill
      \includegraphics[width=0.19\linewidth]{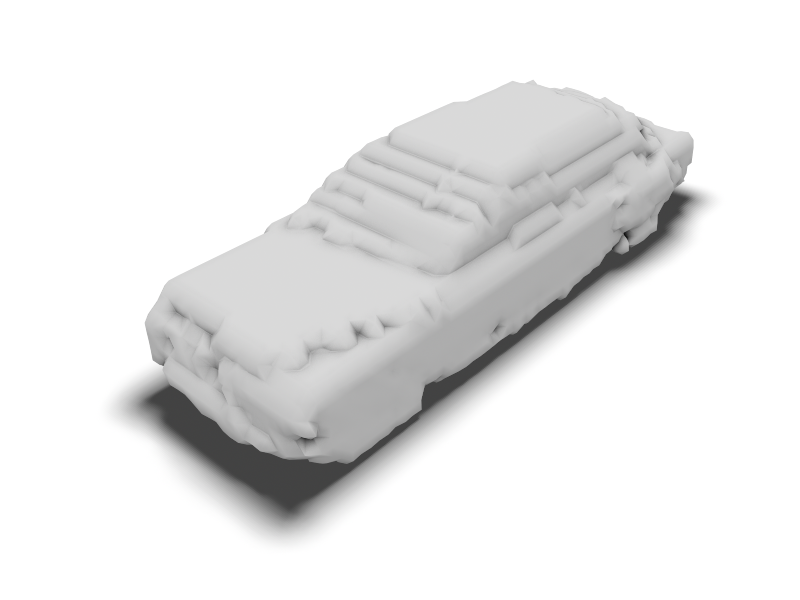}
      \includegraphics[width=0.19\linewidth]{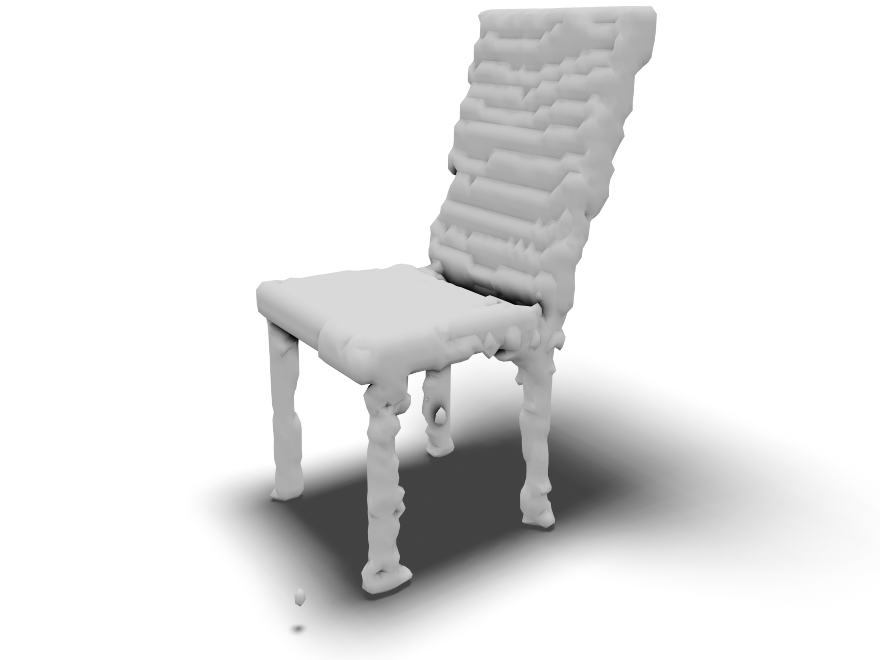}
      \includegraphics[width=0.19\linewidth]{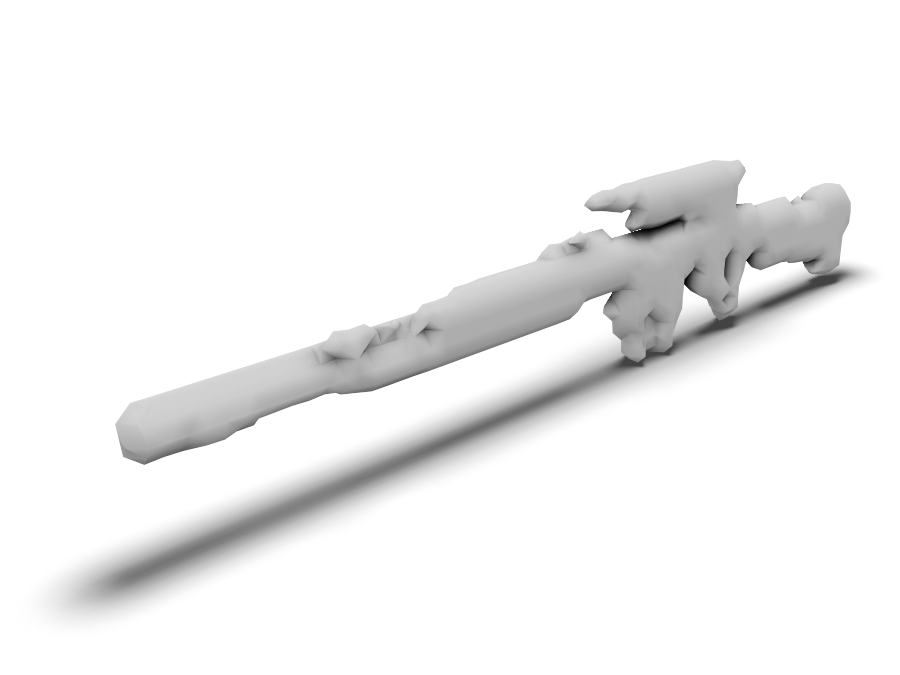}
      \includegraphics[width=0.19\linewidth]{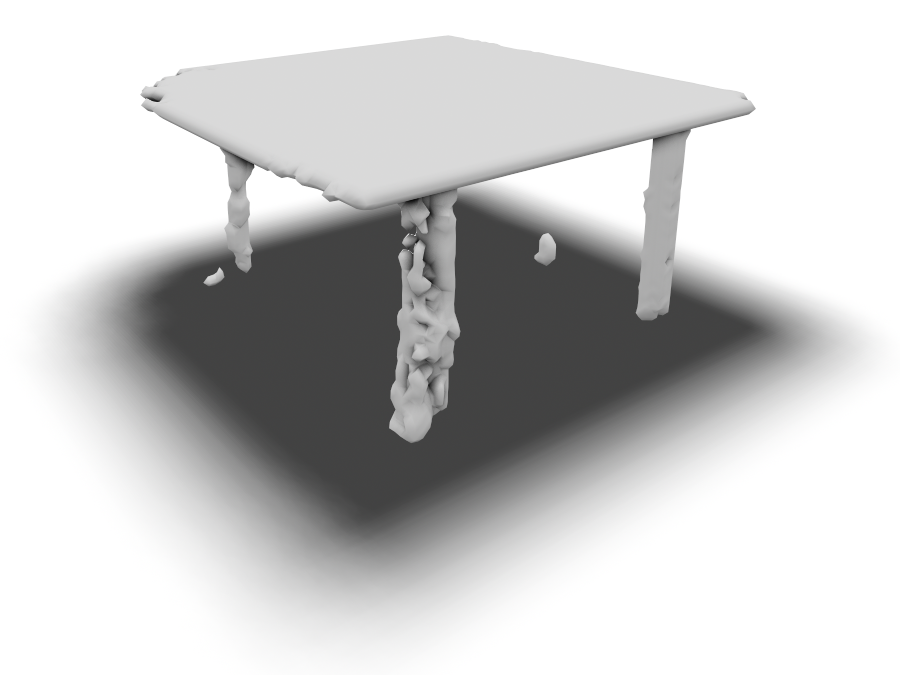}
      \caption{3DGAN}
  \end{subfigure}
  \begin{subfigure}{\linewidth}
      \centering
      \includegraphics[width=0.19\linewidth]{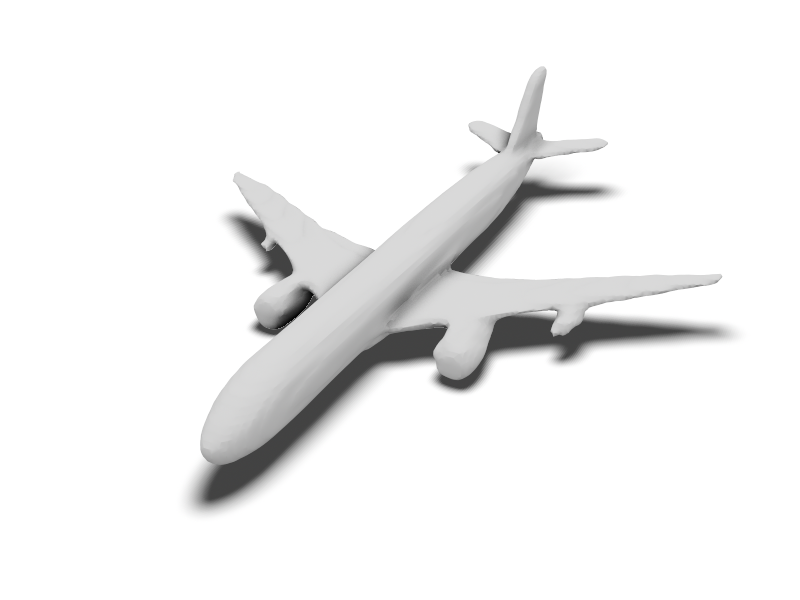}
      \includegraphics[width=0.19\linewidth]{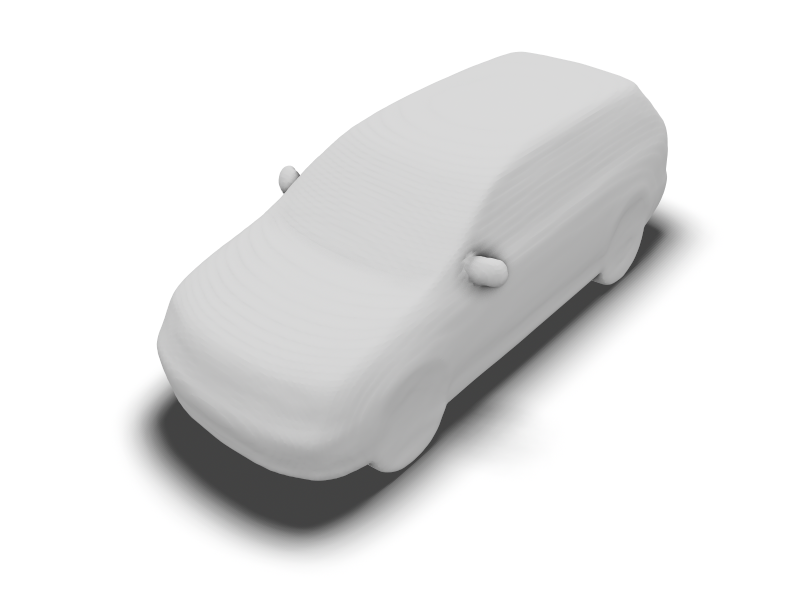}
      \includegraphics[width=0.19\linewidth]{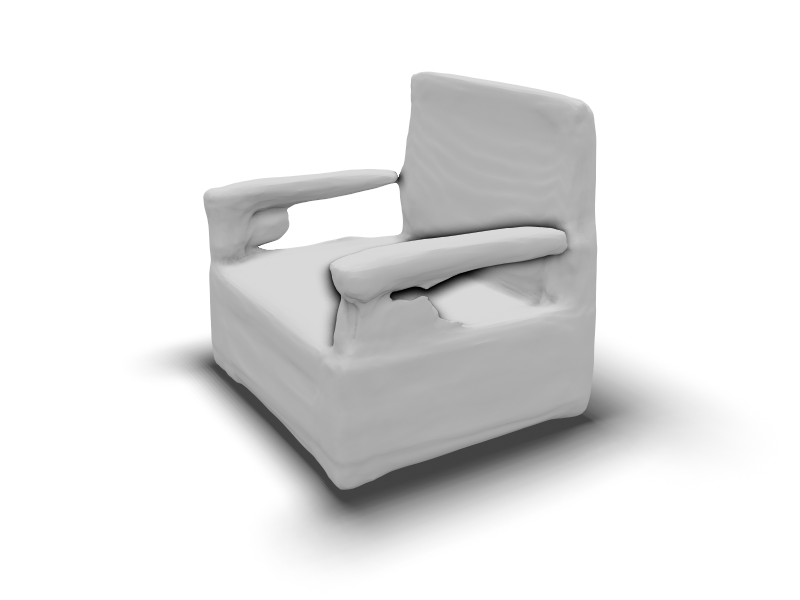}
      \includegraphics[width=0.19\linewidth]{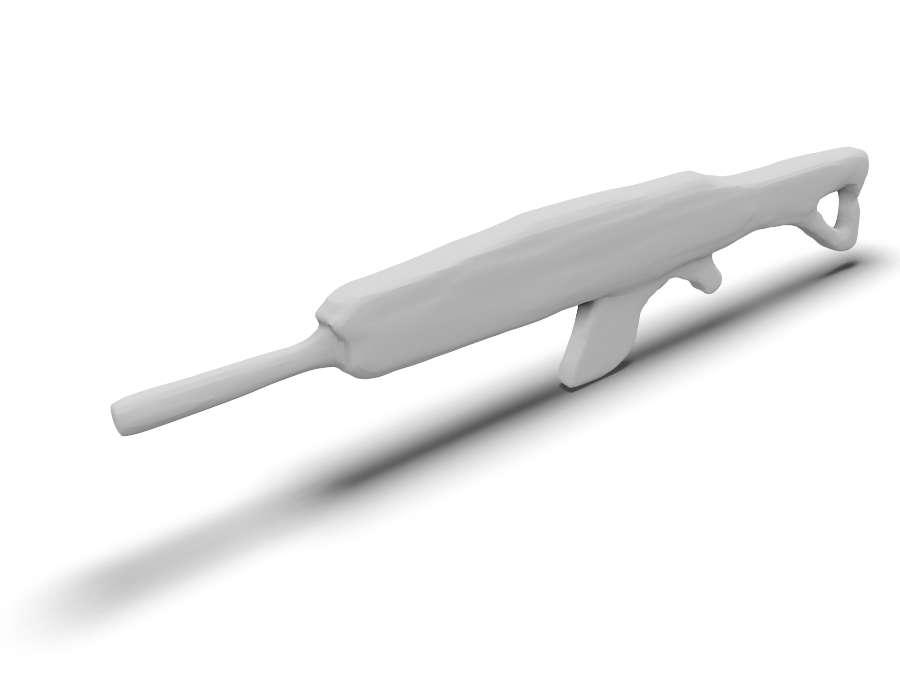}
      \includegraphics[width=0.19\linewidth]{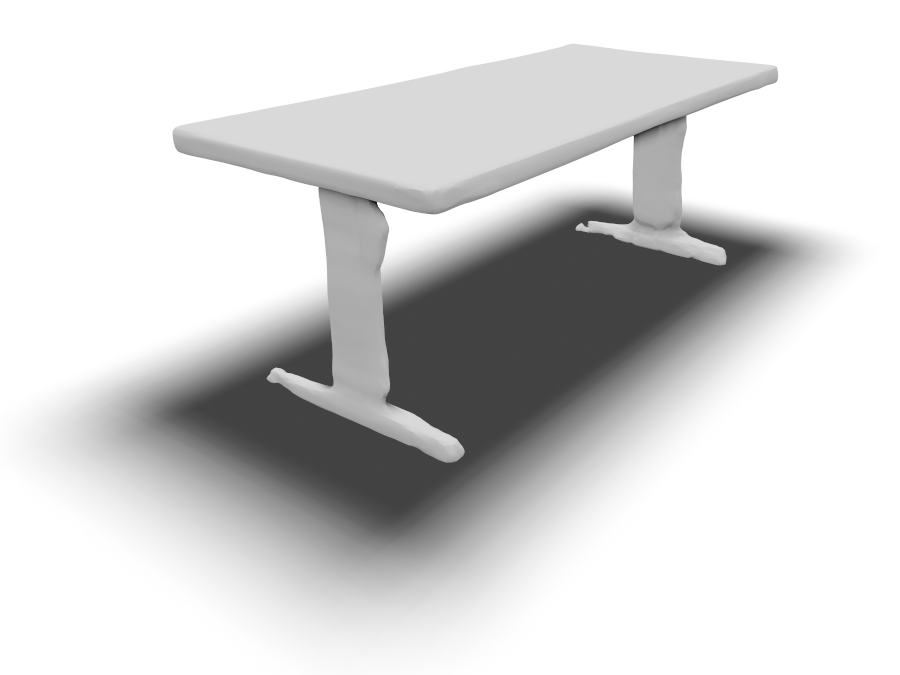}
      \caption{IM-GAN}
  \end{subfigure}
  \begin{subfigure}{\linewidth}
      \centering
      \includegraphics[width=0.19\linewidth]{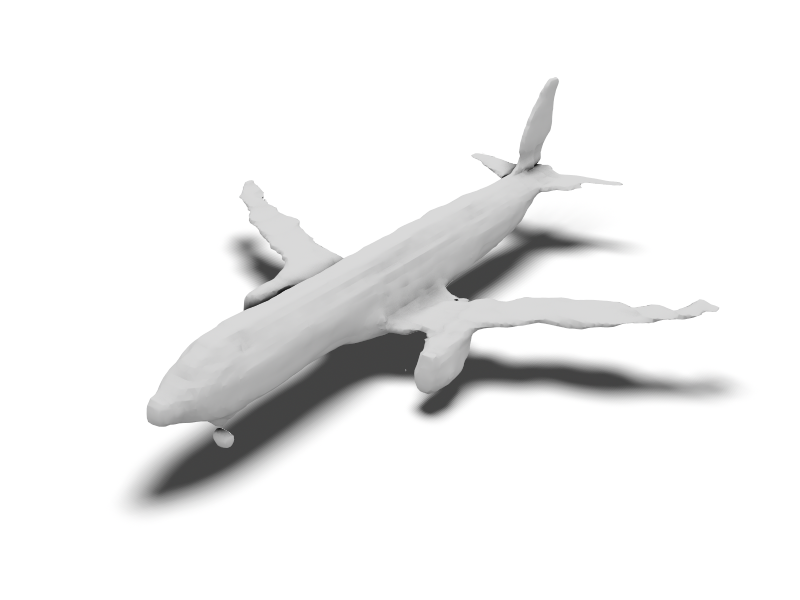}
      \includegraphics[width=0.19\linewidth]{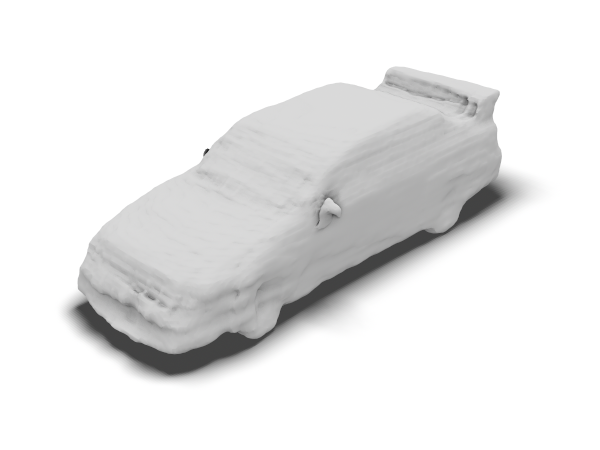}
      \includegraphics[width=0.19\linewidth]{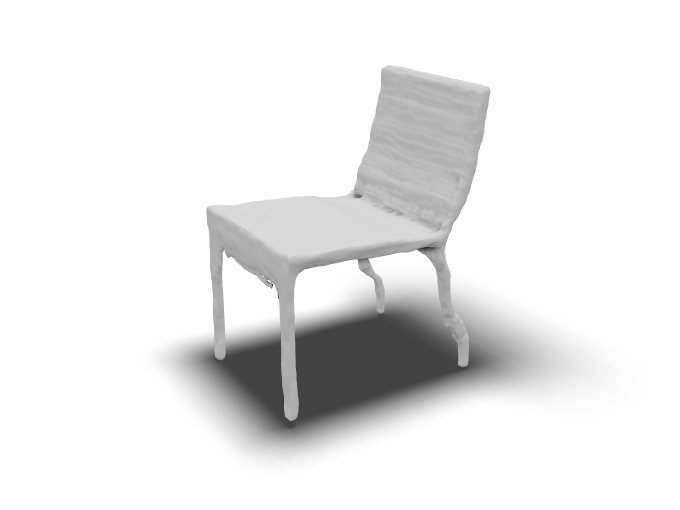}
      \includegraphics[width=0.19\linewidth]{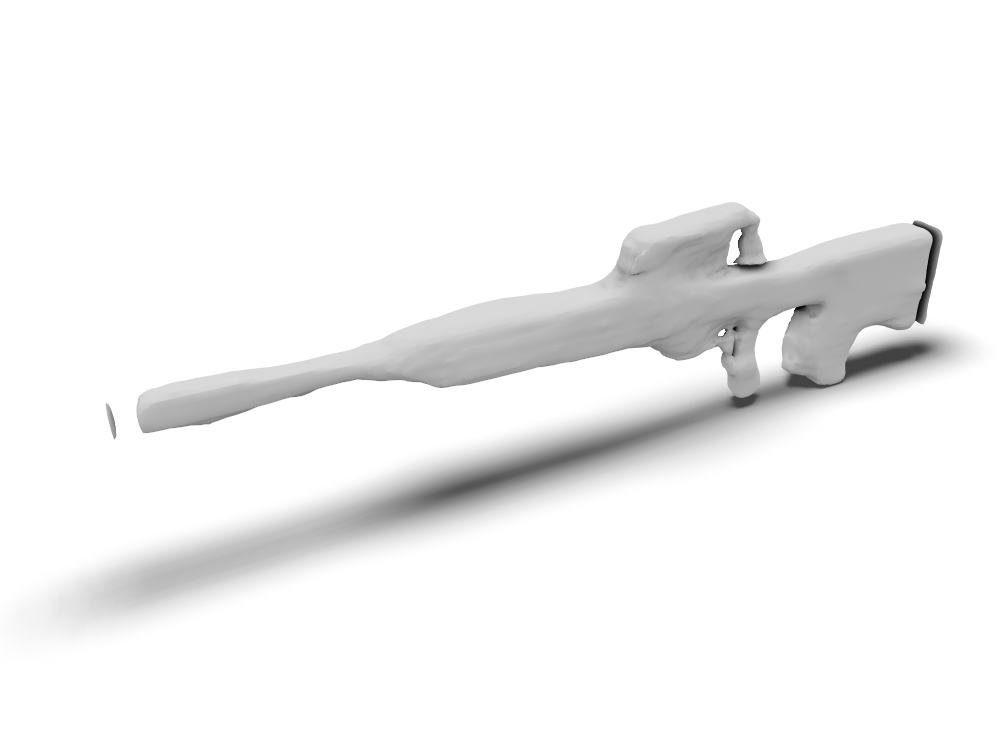}
      \includegraphics[width=0.19\linewidth]{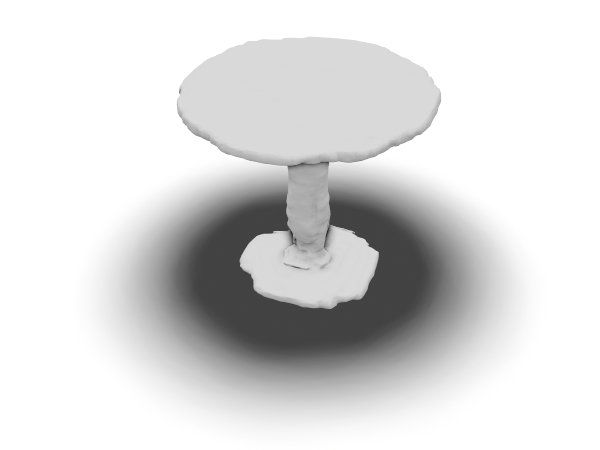}
      \caption{our results}
  \end{subfigure}
  \caption{Typical generative results for each model on each category. For 3DGAN we do not show a plane model, since it was not trained on the plane dataset.}
  \label{fig:results}
\end{figure*}

\begin{figure}
  \centering
  \includegraphics[width=0.49\linewidth]{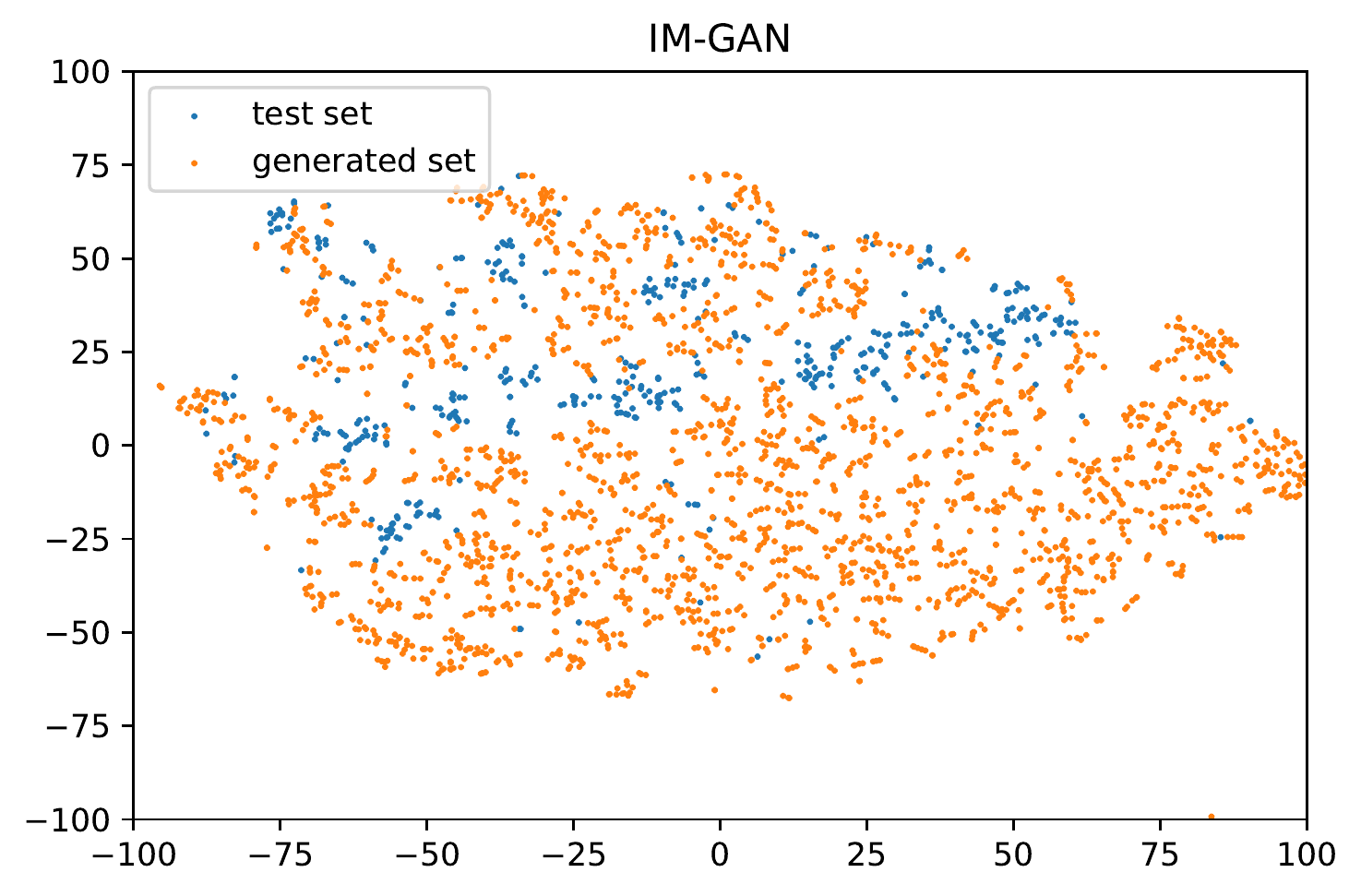}
  \includegraphics[width=0.49\linewidth]{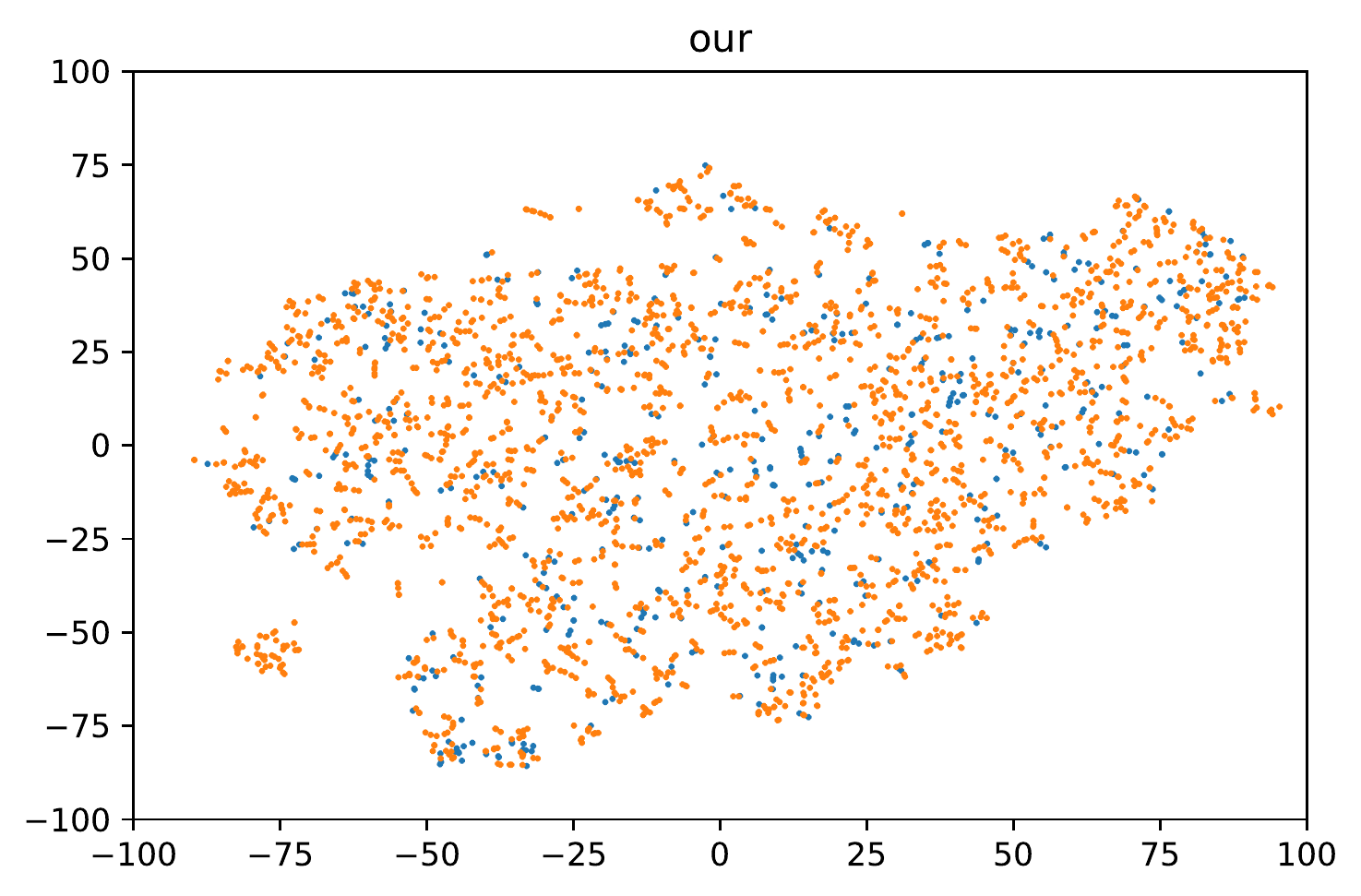}
  \caption{t-SNE~\cite{maaten2008visualizing} (perplexity = 10) low dimensional embeddings of the combined generated and test set from the rifle dataset. On the left distinct clusters are visible, hinting at different densities of the sets. This is not the case on the right.}
  \label{fig:bias}
\end{figure}

To compare our method to the state of the art, we evaluate the results w.r.t. both previous measures Coverage and MMD (Table~\ref{results-table}). We compare to 3DGAN \cite{wu2016learning} (a voxel based method), IM-GAN \cite{chen2019implicit_decoder}, which is a latent-GAN that works on global implicit functions, as well as PC-GAN \cite{Achlioptas2018LearningRA}. Note that a comparison here is difficult, since this method produces point clouds that need to be transformed to meshes first. For all other methods, meshes are created from generated or sampled voxel grids with marching cubes \cite{lorensen1987marching}. The quantitative results for previous methods are taken from \cite{chen2019implicit_decoder}.
Furthermore, following~\cite{chen2019implicit_decoder} all measures are evaluated on meshes that have been extracted from a voxelisation with resolution $64^3$. Therefore, the different methods have been trained and sampled at resolution $64^3$ as well. The size of the generated dataset is always 5 times the size of the test dataset. Note that while IM-GAN and PC-GAN trained specific AEs for each separate class, we trained a single AE on the train split of the entire shapenet dataset. Table~\ref{results-table} shows that our method outperforms the state of the art when it comes to coverage on all but one class, showing the higher diversity of our generated samples. Furthermore, we always have a lower MMD showcasing the higher quality of our sampled shapes.

We also compare against 3DGAN and IM-GAN on the introduced ECD (Table. \ref{ecd-table})
Here we average the score of 10 random sub-samplings of the generated set with equal number to the test set. Qualitative results of the different methods can be seen in Figure \ref{fig:results}. Although the results of IM-GAN sometimes appear smoother than the ones our method produces, this comes at the cost of less diversity.

As we can see in Figure \ref{fig:bias} on the example of the rifle dataset, IM-GAN does not fully capture the distribution of the test set, whereas for our method both distributions are well mixed. This is supported by our quantitative evaluation, where we strongly outperform IM-GAN on the ECD even though our qualitative results are comparable.

A reason for the weaker quantitative results of IM-GAN might be that due to problems during GAN training we are observing mode collapse.
Another possible explanation for this effect might be found with the strong shape bias the AE enforces, as this bias can make it harder for the GAN to faithfully capture the data distribution. This could be due to the AE mapping larger regions of the latent space to similar shapes.
The localized latent grid structure we employ is less prone to such problems, as it is able to represent a much wider variety of shapes.

\paragraph{Conditional Generation}%

\begin{figure}
  \centering
  \includegraphics[width=0.49\linewidth]{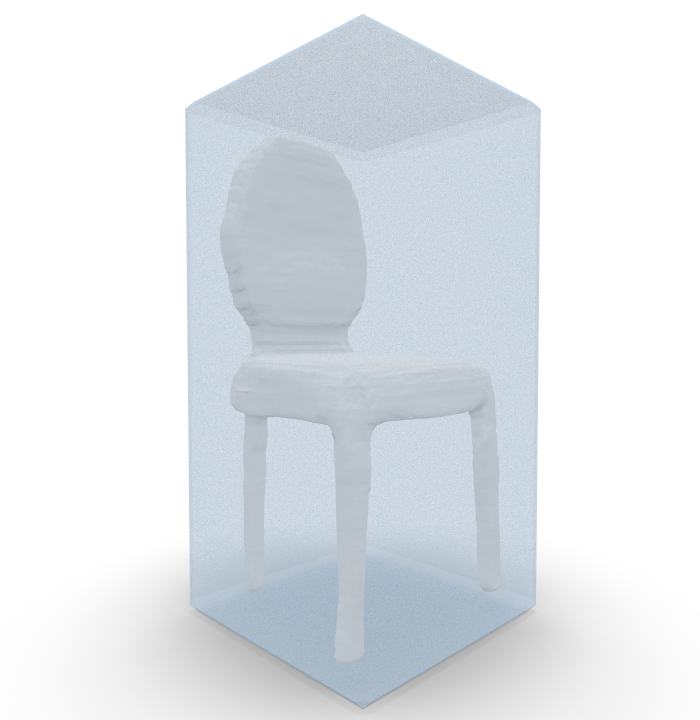}
  \includegraphics[width=0.49\linewidth]{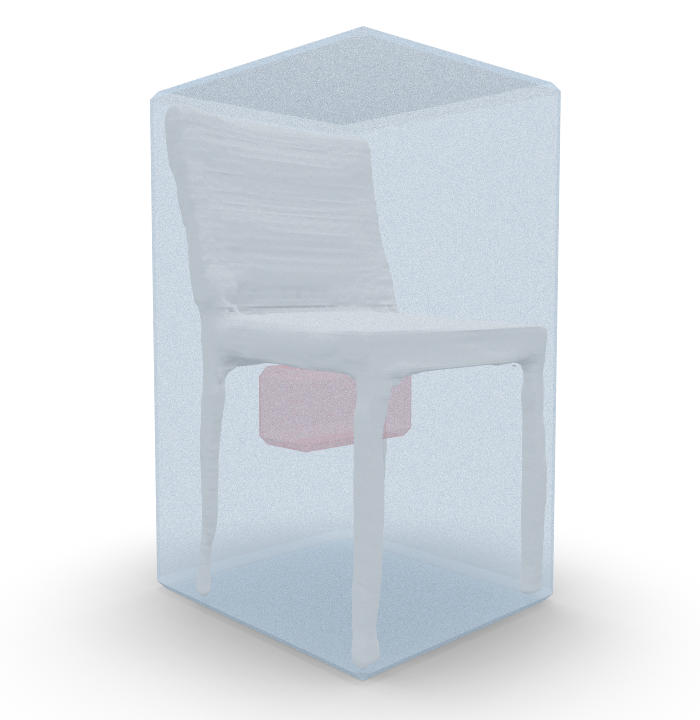}
  \caption{Conditionally generated examples shown with their input bounding boxes. On the right side, we added an additional "negative" box, that is supposed to stay empty.}
  \label{fig:conditional}
\end{figure}

\begin{figure}
  \centering
  \includegraphics[width=0.49\linewidth]{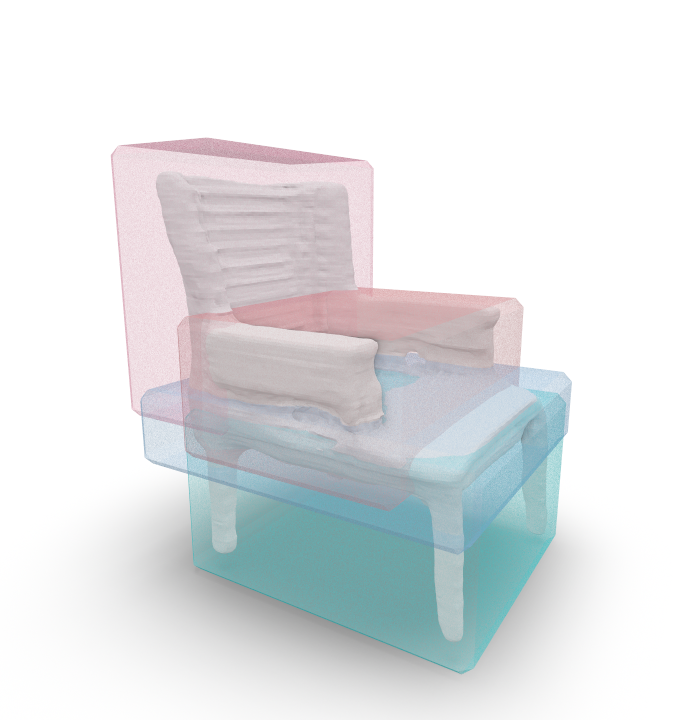}
  \includegraphics[width=0.49\linewidth]{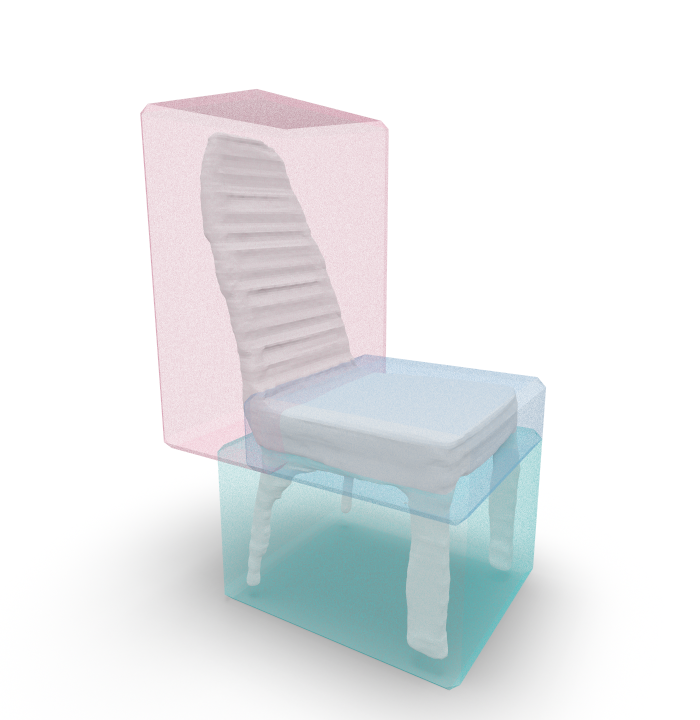}
  \caption{Example for shapes conditioned on part based bounding boxes. The differently colored bounding boxes represent different semantic parts.}
  \label{fig:part_cond}
\end{figure}

As described in Section~\ref{sec:conditional}, our method is able to incorporate spatial guidance into the generation process. As we make use of the grid structure of our latent space for this, approaches based on a global latent space are unable to offer such guidance in a straightforward manner.
This guidance can be provided in many different ways. This information can take the form of binary values, class labels, or even complex encodings. This flexibility allows for several applications, some of which we will present in the following.

The simplest possibility to offer guidance is to provide a single binary variable for each grid cell. With this we can for example generate shapes that fit into a predefined bounding box. This bounding box is simply discretized to our grid resolution so that each cell gets the information whether it is inside or outside of it.
Exemplary results of this generation process, as well as the bounding boxes they were conditioned on, can be seen in Figure \ref{fig:conditional}.
We can not only prescribe a bounding box, that should contain all of the geometry, but allow the user to further restrict the shape space, by specifying ``negative'' bounding boxes, that should not contain any geometry.

If semantic information is available (in our case given by the PartNet dataset \cite{mo2019CVPR}), this can be used for more fine grained generation (Figure \ref{fig:part_cond}). This means we can prescribe a bounding box for each semantic part of the object. In this case, we do not provide binary data, but part labels per grid cell.

\begin{figure}
  \centering
  \begin{overpic}[width=0.49\linewidth]{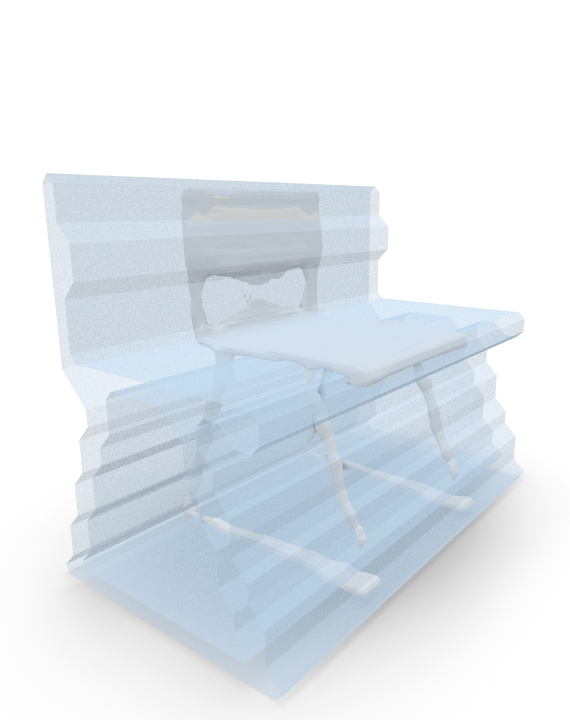}
     \put(3,3){\includegraphics[width=0.1\linewidth]{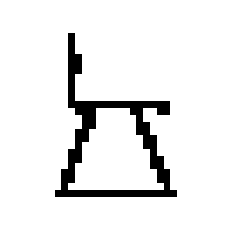}}  
  \end{overpic}
  \begin{overpic}[width=0.49\linewidth]{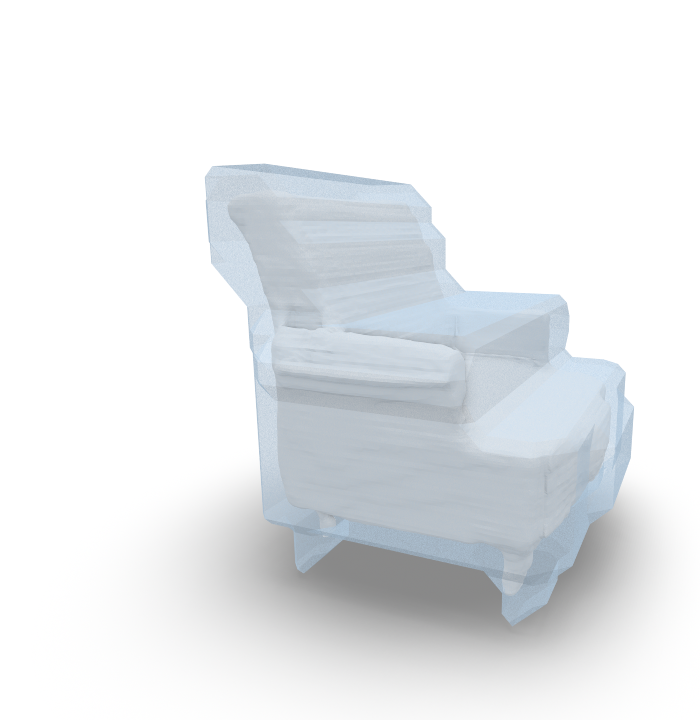}
     \put(3,3){\includegraphics[width=0.1\linewidth]{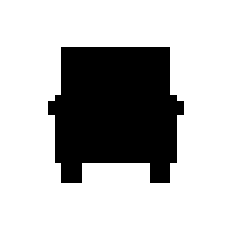}} 
     \put(3,25){\includegraphics[width=0.1\linewidth]{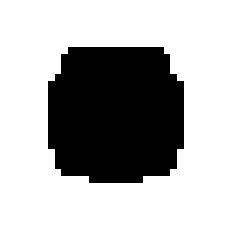}}  
     \put(3,50){\includegraphics[width=0.1\linewidth]{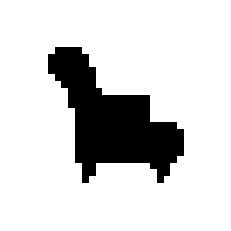}}   
  \end{overpic}
  \caption{Example for silhouette based generation. The 3D masks are extracted from the low resolution silhouettes shown in the insets.}
  \label{fig:silhouette}
\end{figure}

Bounding boxes are however not the only option to provide this simple guidance. Another possible application, would be to compute this information from a silhouette, given as a binary 2D image (Figure \ref{fig:silhouette}). For simplicity, we assume the viewing direction to be axis aligned. The information can be presented as a single image, or as multiple images corresponding to different axes.

\begin{figure}
  \centering
  \includegraphics[width=\linewidth]{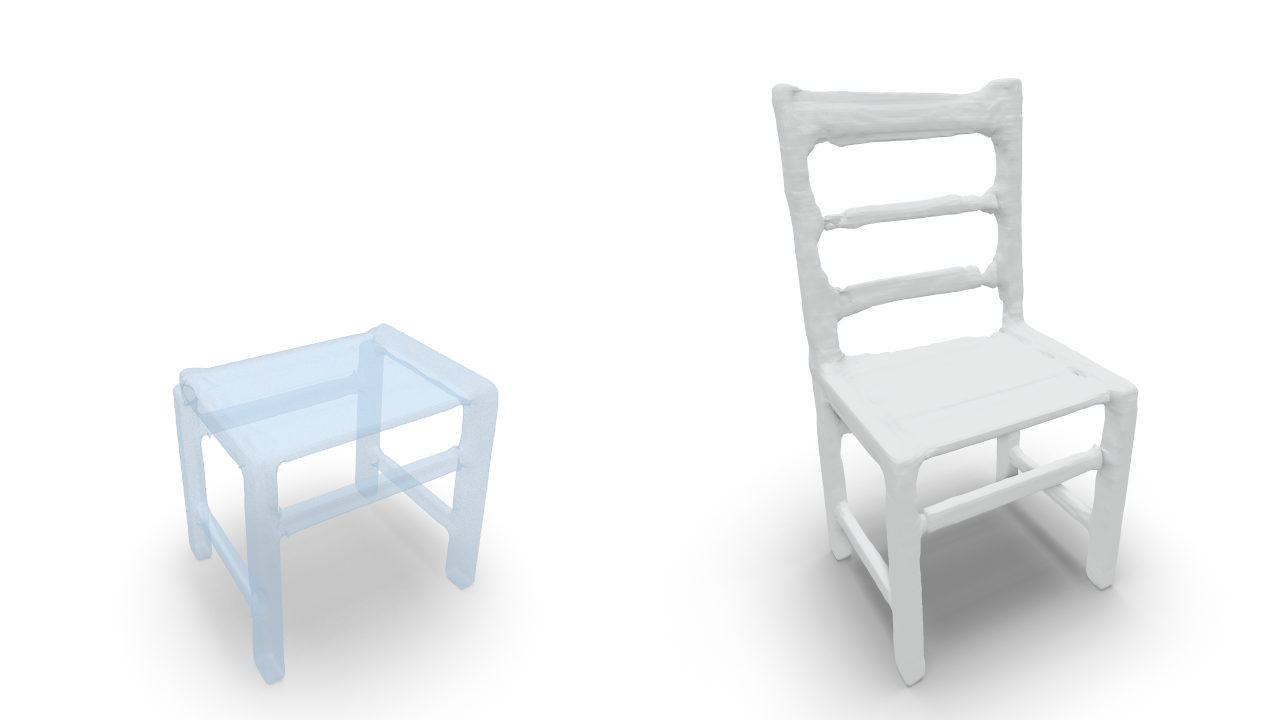}
  \caption{Example for shape completion. On the left we see the given conditioning, on the right the generated chair.}
  \label{fig:completion}
\end{figure}

In all previous examples, we provided one-hot vectors per cell. However, complex encodings can be used for this kind of conditioning as well. As an example, we show the application of shape completion on chairs. For this we cut out one semantic part from the object (backrest, armrest, seat or legs) and train the network to insert this missing part. This is implemented by conditioning the GAN on the geometry that is still available. We provide this information by using our AE to encode the present geometry into latent vectors for each cell (Figure \ref{fig:completion})
As can be seen, our generated shape follows the provided guidance, where it is given, and is reasonably completed, where it is not.
For all examples we obtained the supervision masks from unseen examples of the test set.

\section{Conclusion}
We introduced a GAN that generates piecewise implicit functions organized in grids to represent 3D shapes. By learning on localized latent representations instead of global ones (as in previous work) we are able to model the data generating distribution more closely than prior methods. We showed this by evaluating our method quantitatively with measures from prior work as well as with the proposed usage of ECD. Due to the convolutional nature of our GAN architecture we are able to incorporate spatial guidance (in a wide variety of forms) in the generation process. We only showed few examples, but expect this approach to viable for a wide range of tasks.

The results generated from global latent representations appear smoother since they are less likely to capture the details of the ground truth.
As the AE part in our generation process is able to learn fine details of the underlying shapes, it tends to reproduce voxelization artifacts from the ground truth.
We therefore expect higher resolution ground truth shapes to lead to even better results for our method.

\paragraph*{Acknowledgements}
This work was supported by the Gottfried-Wilhelm-Leibniz Programme of the Deutsche Forschungsgemeinschaft DFG, project number KO2064/6-1, as well as DFG project number KO2064/9-1

\bibliographystyle{abbrv}
\bibliography{main}    

\include{supplementary}
\end{document}

%% file: supplementary.tex
\begin{appendices}

\section{Edge Count Difference}\label{sec:ecd}

The exact formula to compute the Edge Count Difference (ECD)~\cite{chen2017new} between sets $\mathcal{A}$ and $\mathcal{B}$, used as evaluation measure for all experiments.

\begin{equation}
    \text{ECD}=(R_1 - \mu_1, R_2 -\mu_2) \Sigma^{-1} 
    \begin{pmatrix} {R_1 - \mu_1} \\ {R_2 - \mu_2}\end{pmatrix}
\end{equation}

$R_1$ and $R_2$ are the counted edges within $\mathcal{A}$ and $\mathcal{B}$ respectively.
The expected values for $R_1$ and $R_2$ are given as $\mu_1$ and $\mu_2$.
$\Sigma$ is the covariance matrix of the vector $(R_1, R_2)$ under the permutation null distribution.
Specifically, that means
\begin{align*}
    \mu_1 &= |G|\frac{n(n-1)}{N(N-1)}\\
    \mu_2 &= |G|\frac{m(m-1)}{N(N-1)}\\
    \Sigma_{11} &= \mu_1(1-\mu_1) + 2C \frac{n(n-1)(n-2)}{N(N-1)(N-2)} \\
    & +(|G|(|G|-1) - 2C) \frac{n(n-2)(n-2)(n-3)}{N(N-1)(N-2)(N-3)}\\
    \Sigma_{22} &= \mu_2(1-\mu_2) + 2C \frac{m(m-1)(m-2)}{N(N-1)(N-2)} \\
    & +(|G|(|G|-1) - 2C) \frac{m(m-2)(m-2)(m-3)}{N(N-1)(N-2)(N-3)}\\
    \Sigma_{12} &= \Sigma_{21} = (|G|(|G|-1)-2C) \\
    & \cdot \frac{nm(n-1)(m-1)}{N(N-1)(N-2)(N-3)} - \mu_1 \mu_2
\end{align*}

where $G$ is the k-MST build from set $\mathcal{A}$ and $\mathcal{B}$, $n = |\mathcal{A}|$, m = $|\mathcal{B}|$ and $N = m+n$. C is given as $C = \frac{1}{2} \sum^N_{i=1} |G_i|^2 - |G|$, with $G_i$ being the subgraph in G that includes all edges that connect to node $i$. All formulas are from \cite{chen2017new}.
\begin{wrapfigure}{r}{0.5\linewidth}
  \begin{center}
    \includegraphics[width=\linewidth]{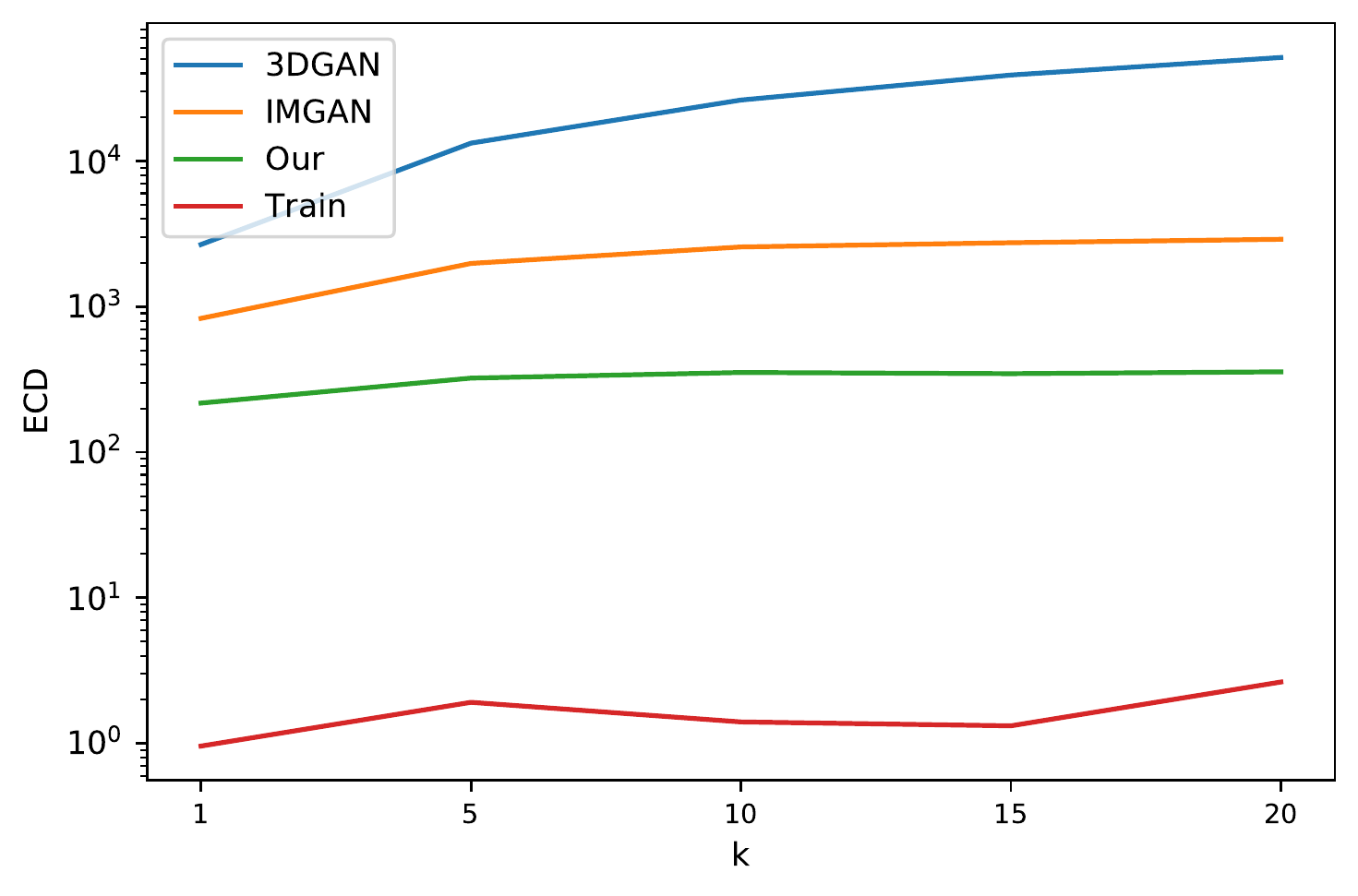}
  \end{center}
\end{wrapfigure}
The only parameter of this method is $k$, the minimal number of neighbors of each vertex in the MST. Although this parameter does effect the magnitude of the score, relative distances do not change significantly, as we show on the right on the example of the chair dataset. For all of our experiments $k$ is set to 10.

\section{Network Architectures} \label{sec:arch}

In this section we go into detail on all architectures used in our experiments. In our figures a rectangle signifies data, with its given size. A rounded rectangle stands for layers of our networks. For convolutions we note the number of channels, the kernel size and the stride.

\vspace{2cm}

\begin{figure}[htb]
  \centering
  \includegraphics[scale=0.5]{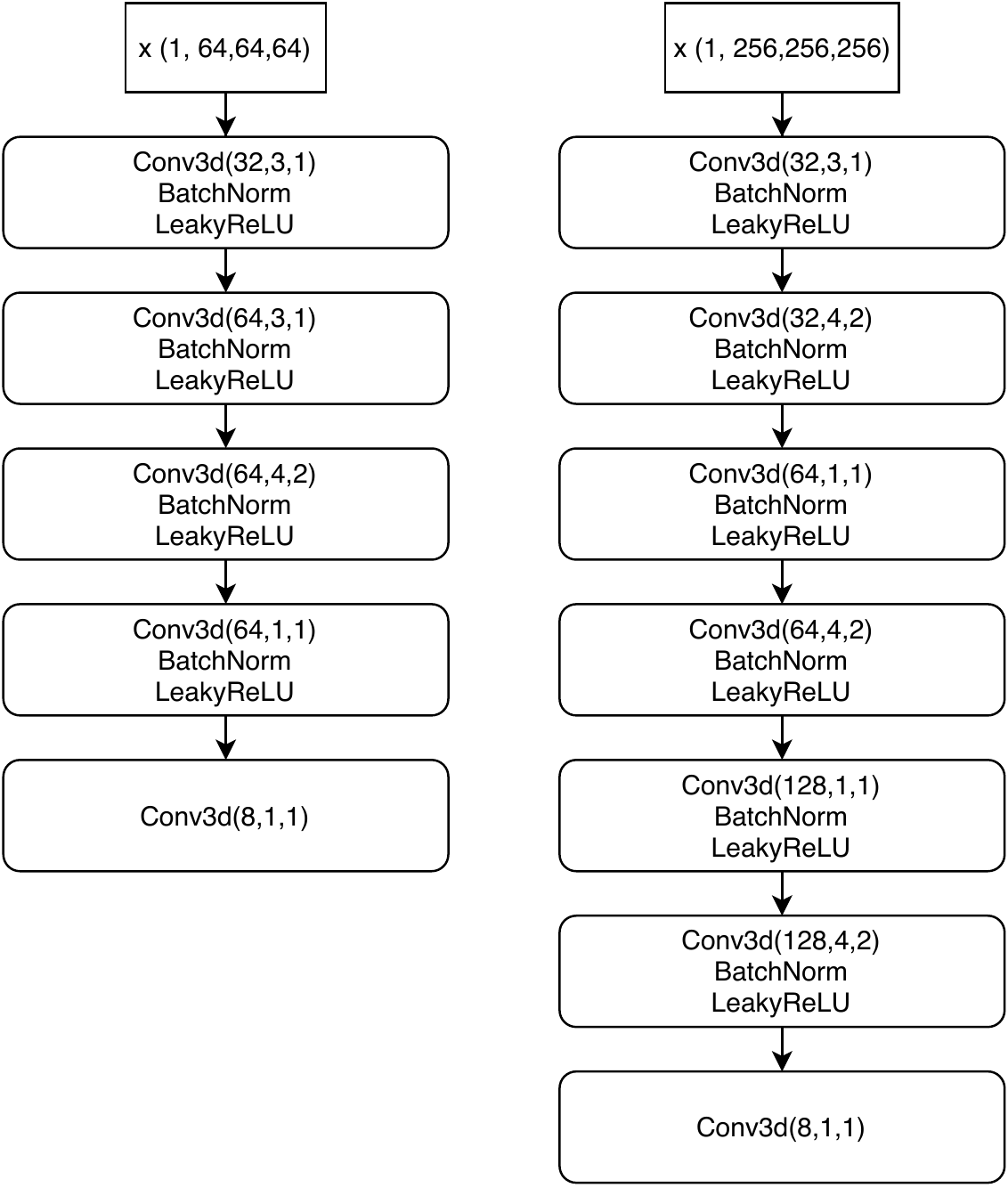}
  \caption{Two different encoders used for high or low input resolutions. In both cases the same decoder is used. The encoders receive as input a voxel grid with a resolution of 64 or 256 respectively. In both cases the output resolution is 32 and 8 channels are used}
  \label{fig:encoder}
\end{figure}

\begin{figure}[htb]
  \centering
  \includegraphics[scale=0.5]{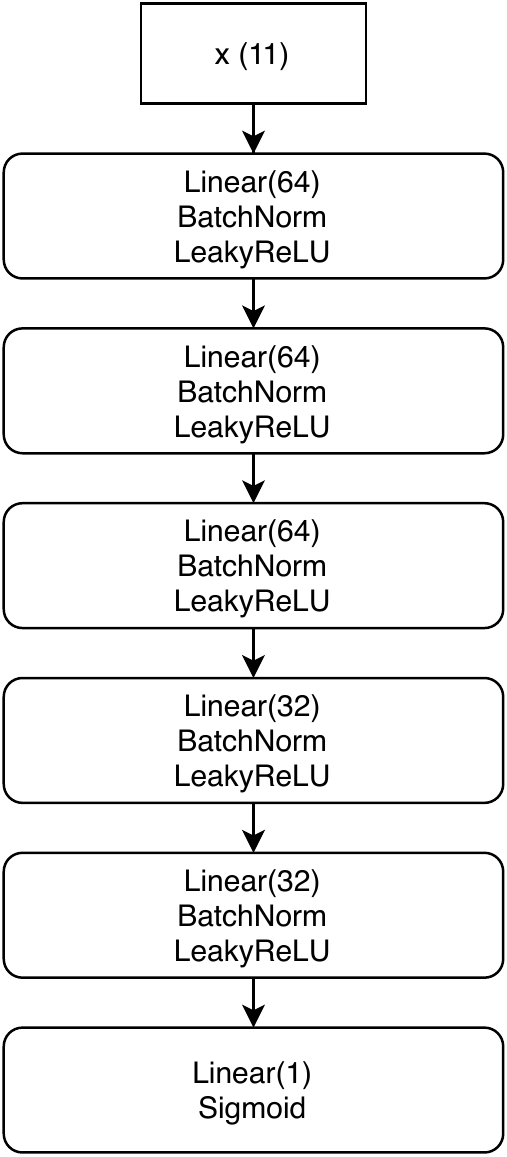}
  \caption{The decoder used in all our experiments. It gets as input a points coordinates relative to the cell center it is located in, concatenated with the cells latent vector. The output can be rounded to a binary value, telling us whether the point is inside our outside of the shape}
  \label{fig:decoder}
\end{figure}

\begin{figure}[htb]
  \centering
  \includegraphics[scale=0.5]{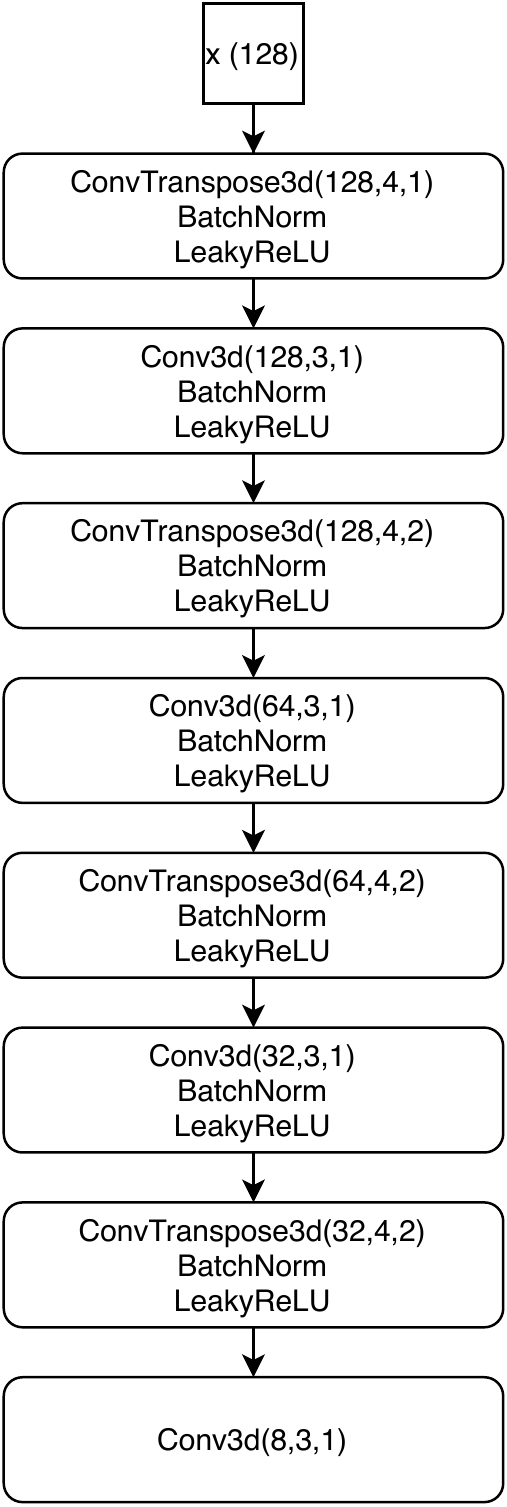}
  \caption{For unconditional generation we use a generator with a simple convolutional architecture.}
  \label{fig:generator}
\end{figure}

\begin{figure}[htb]
  \centering
  \includegraphics[scale=0.5]{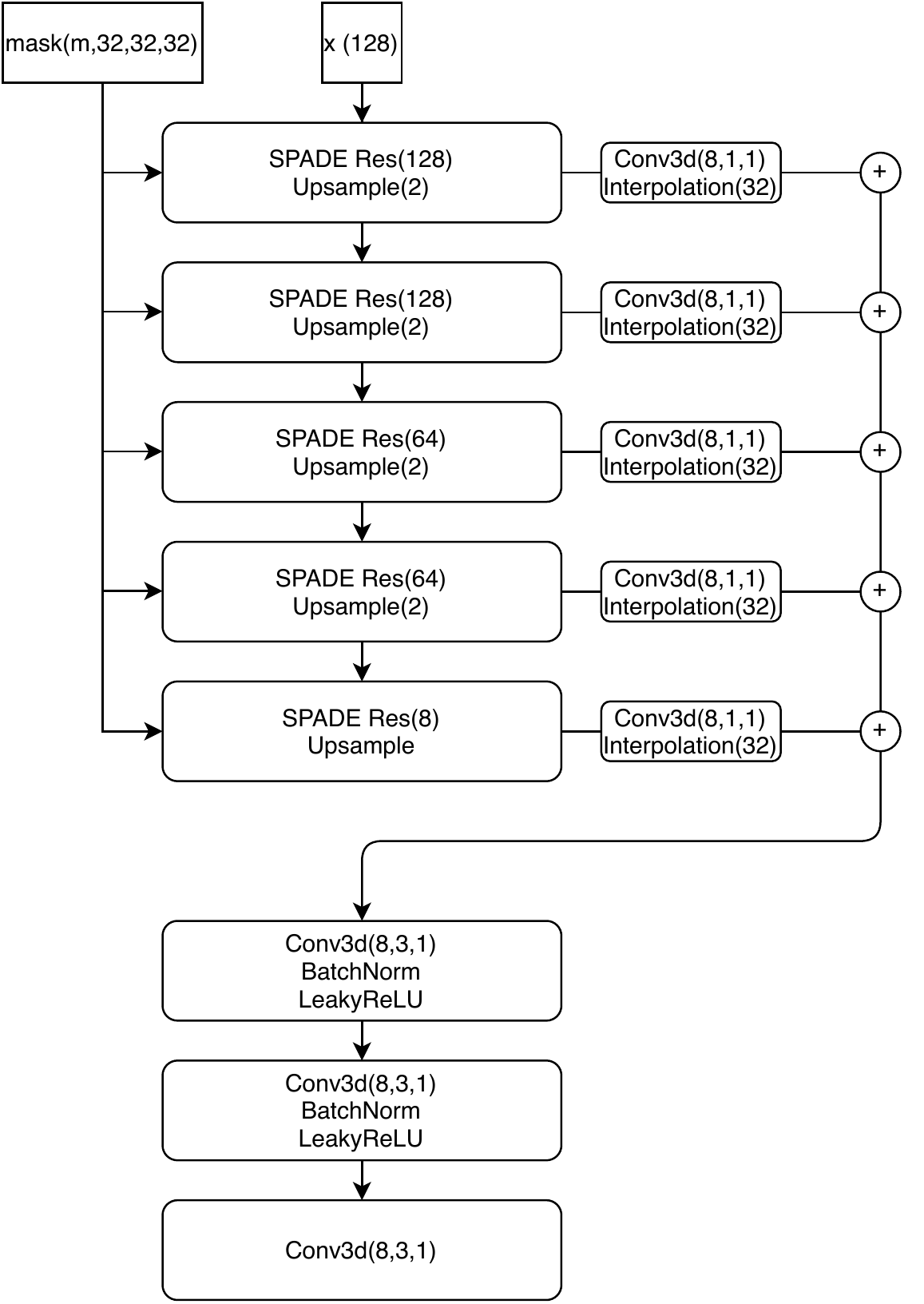}
  \caption{Our conditional generator is inspired by SPADE \cite{park2019semantic}, where the mask is used to compute cell-wise scales and biases. The number of channels $m$ depends on the application. For the design of the SPADE Res block we refer to \cite{park2019semantic}. We added skip connections to their architecture.}
  \label{fig:generator-cond}
\end{figure}

\begin{figure}[htb]
  \centering
  \includegraphics[scale=0.5]{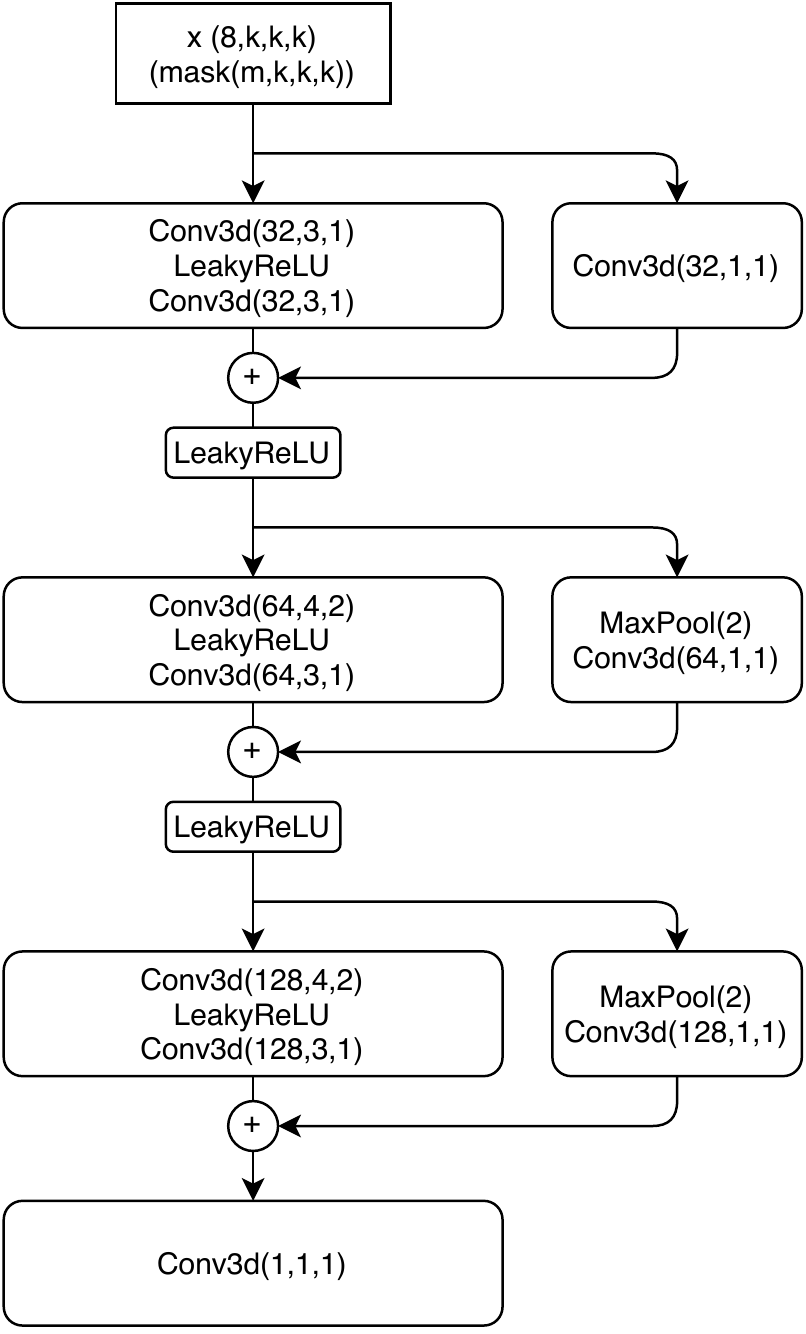}
  \caption{The patch discriminator used in all experiments. Spectral normalization is applied to all convolutional layers. The input resolution $k$ is either 32, 16 or 8. For conditional generation the number of mask channels $m$ depends on the application. The mask is simply concatenated to the latent grid.}
  \label{fig:discriminator}
\end{figure}

\section{Training} \label{sec:training}
All experiments were done on a GeForce RTX 2080 Ti. We used Adam \cite{kingma2014adam} as optimizer and a learning rate of $1e-3$ for the autoencoder, generator and discriminator with no weight decay.
\paragraph{Autoencoder}
We trained the autoencoder for 200 epochs, although a lower number would probably suffice, as the AE converges fast to satisfactory results. For the high-resolution version we used a batch size of 16 for the low-resolution version of 8. Per object the implicit function was sampled at 6000 positions. For the high-resolution version the entire grid does not fit into memory, therefore we randomly carve a 3D slice of resolution 48 out for processing. 

\paragraph{GAN}
We trained the GAN for 500 epochs with a batch size of 48. The gradient penalty weight was chosen as 1. The training time ranged from 20 hours for unconditional generation on the rifle dataset to 100 hours for conditional generation on the table dataset.

\section{Evaluation} \label{sec:evaluation}

We conducted an ablation study, to show the effect of different choices regarding the discriminator (Table \ref{ablation}). When choosing a regular discriminator, instead of a patch-based architecture, we observe significant mode collapse. Furthermore, we show that each of the three used discriminators improves the training result. It should come as no surprise, that the discriminators at higher resolution are more important for the results.
\begin{table}[htb]
  \centering
  \begin{tabular}{ccccccccc}
    \hline
                & ECD   & MMD   & COV   \\
    \hline
        no patch& 27347 & 6818  &  0.00 \\
        16 8    & 6158  & 3265  & 74.71 \\
        32 8    & 323   & 2778  & 76.40 \\
        32 16   & 180   & 2784  & 81.49 \\
        complete& 144   & 2768  & 82.09 \\
    \hline
  \end{tabular}
  \caption{Ablation study on the chair dataset. We show results for a standard (not patch-based) discriminator. Furthermore, we show, that the results worsens, when leaving out one of the three discriminators. The best results are obtained when using all three.}
  \label{ablation}
\end{table}

As it is straightforward for our method to produce results in higher resolutions, we report numbers at a resolution of 256 as well (Table \ref{results-256}). For these comparisons we do not compute distances to voxelized ground truth meshes but to the original ones. Therefore these numbers are not comparable to our other results, but might be of interest for future comparisons. We furthermore report results for conditional generation. For this we conditioned on the bounding boxes obtained from the test set.

\begin{figure}[htb]
  \centering
  \includegraphics[width=\linewidth]{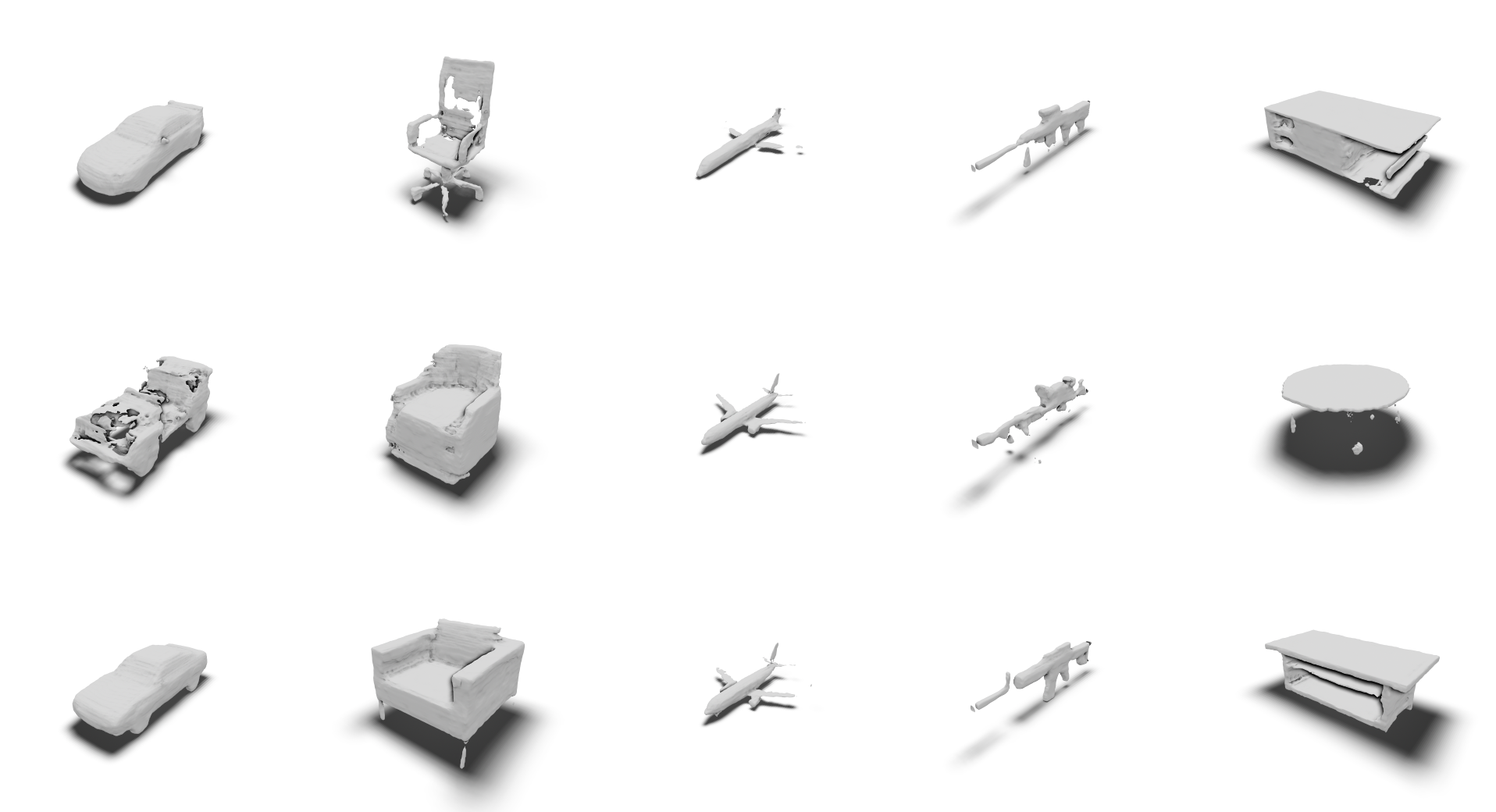}
  \caption{Results from our unconditional generator, sampled at random}
  \label{fig:random_samples}
\end{figure}

\begin{figure}[htb]
  \centering
  \includegraphics[width=\linewidth]{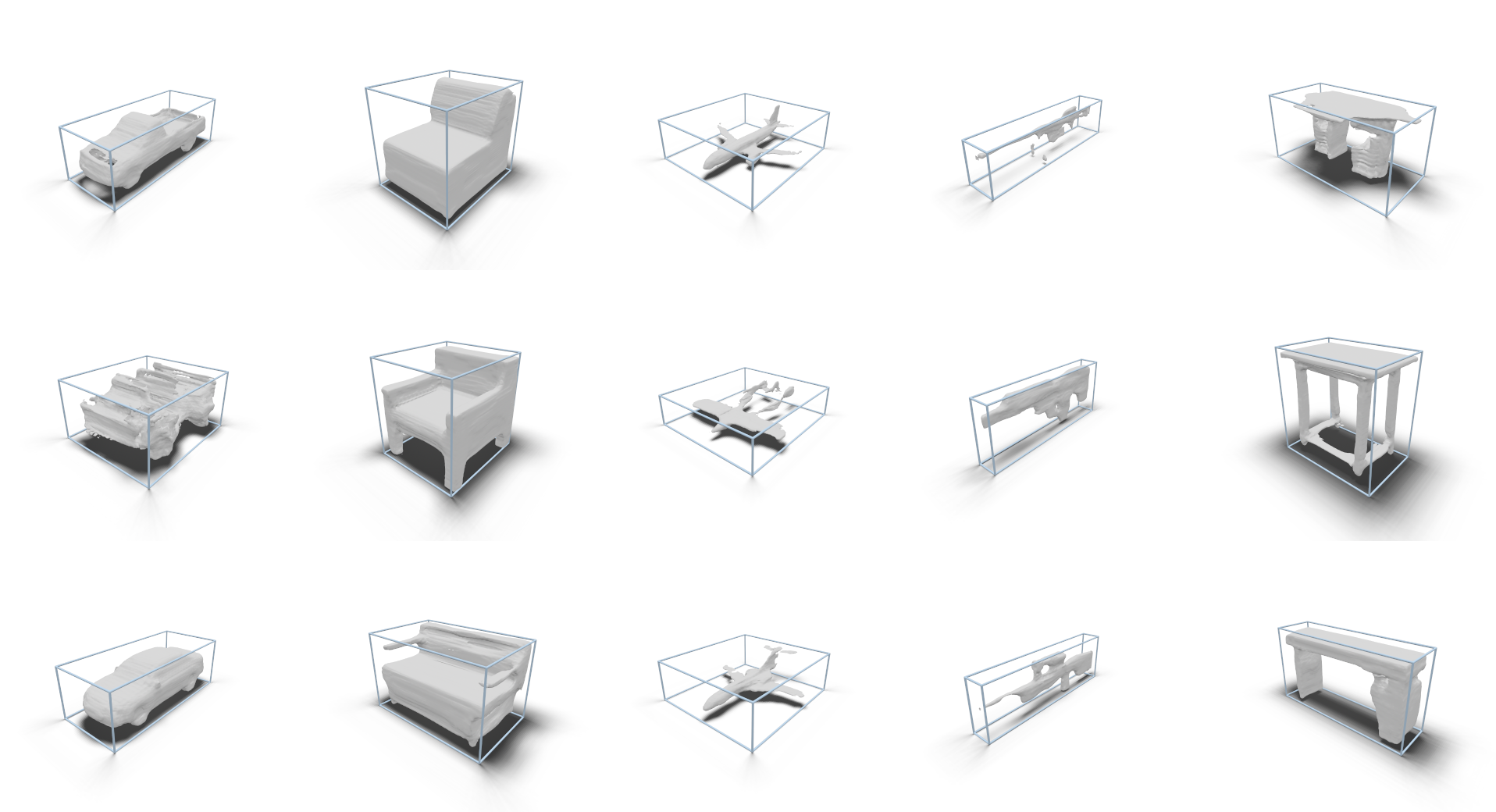}
  \caption{Results from our generator conditioned on bounding boxes, sampled at random}
  \label{fig:random_samples_cond}
\end{figure}

Furthermore, we show additional models generated with our approach both for unconditional (Figure \ref{fig:random_samples}) and bounding box based generation (Figure \ref{fig:random_samples_cond}). The displayed objects are randomly sampled.

\begin{figure}[htb]
  \centering
  \includegraphics[width=0.32\linewidth]{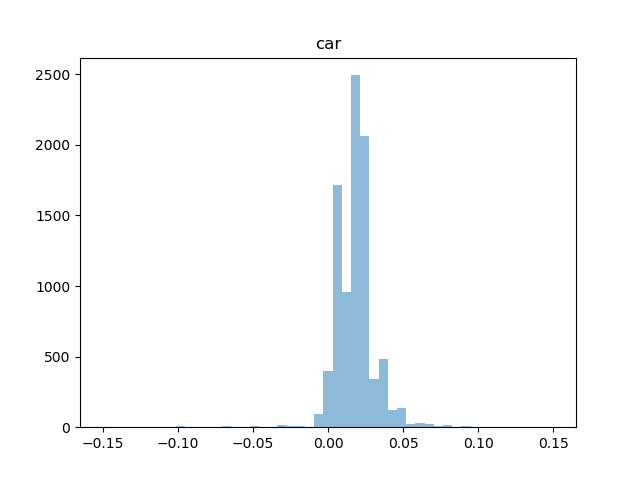}
  \includegraphics[width=0.32\linewidth]{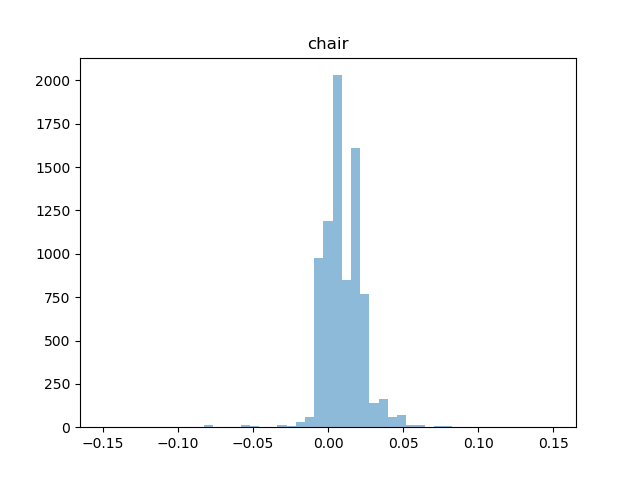}
  \includegraphics[width=0.32\linewidth]{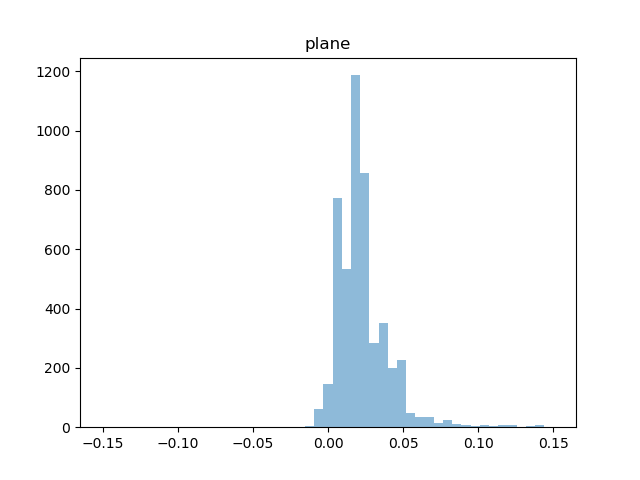}
  \includegraphics[width=0.32\linewidth]{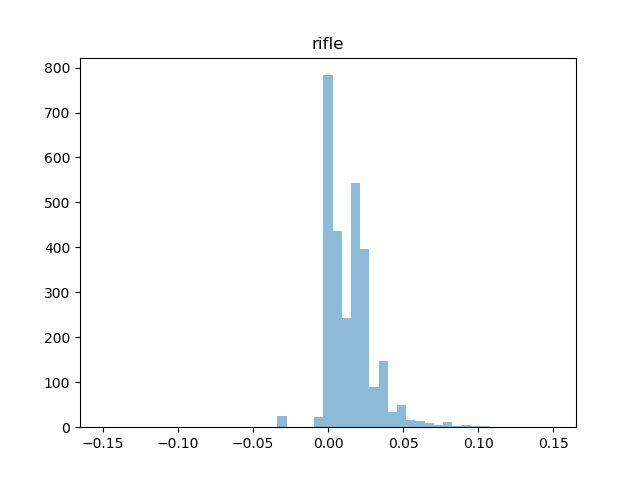}
  \includegraphics[width=0.32\linewidth]{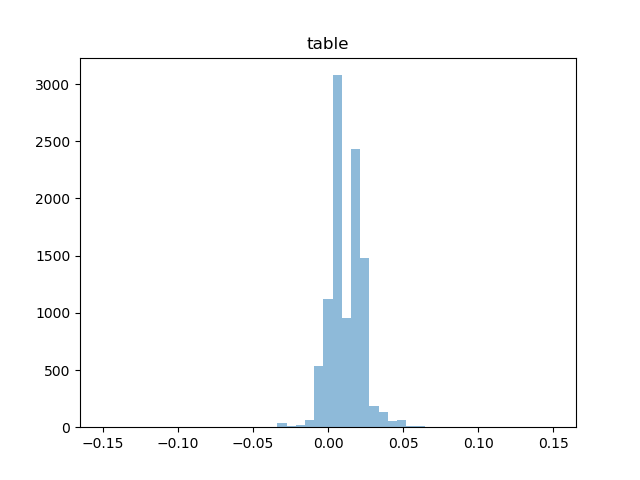}
  \caption{Numeric results to evaluate the fit of the shapes to their bounding boxes. The generation is conditioned on the bounding boxes of the test set. Note that we discretize the bounding boxes when using them as masks. Therefore errors between 0 an 1/32 are expected.}
  \label{fig:conditional-results}
\end{figure}

We further add a numerical evaluation of the bounding box fit. As the bounding box masks are discretized to a resolution of 32, we expect the difference between masks and actual bounding boxes to be between 0 and 1/32. As can be seen in Figure~\ref{fig:conditional-results} the bounding boxes of most of our objects fall into this range.

\begin{figure}[htb]
  \centering
  \includegraphics[width=\linewidth]{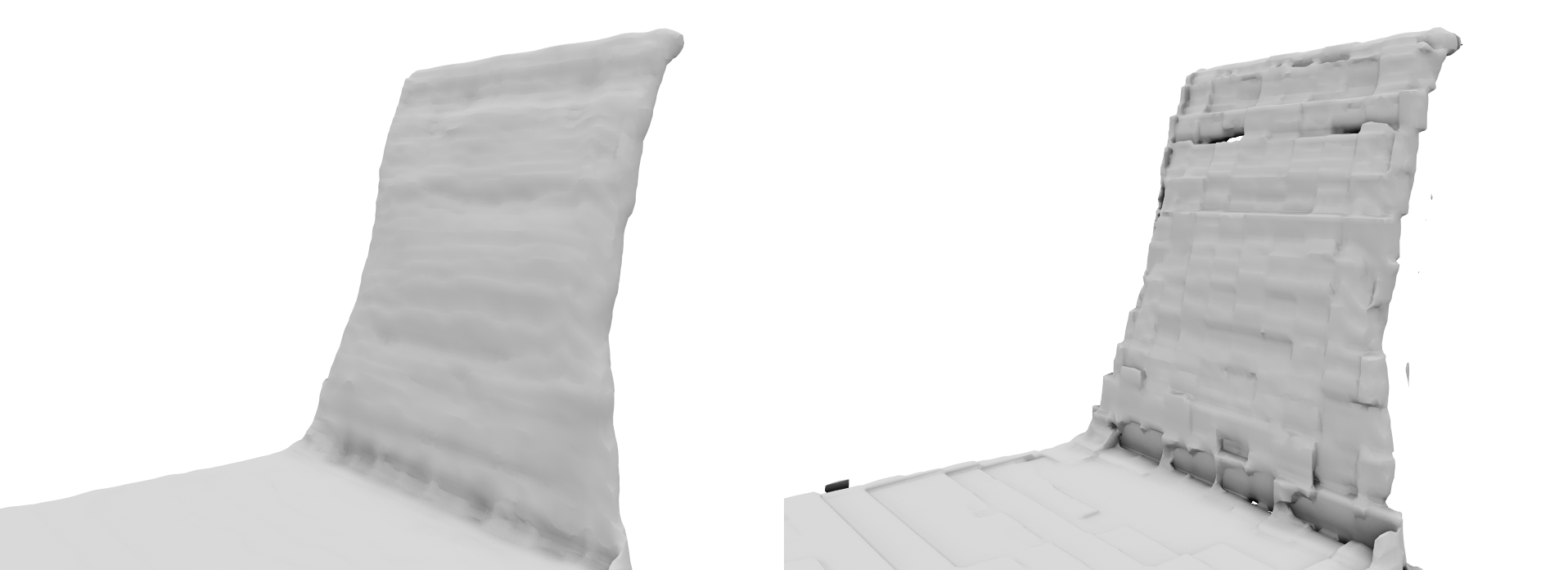}
  \caption{To demonstrate the effect of smoothing the classification results in a trilinear manner we show a generated chair with and without smoothing}
  \label{fig:smoothing}
\end{figure}

Lastly, we demonstrate the effect smoothing has on the autoencoder results (Figure~\ref{fig:smoothing}). When no smoothing is applied distinct borders between individual cells are visible.

\begin{table*}[htb]
  \centering
  \begin{tabular}{ccccccccc}
    \hline
                &               & Plane & Car   & Chair & Rifle & Table &  Avg.  \\
    \hline
        COV(\%) & Unconditional & 76.89 & 74.67 & 82.82 & 73.89 & 85.61 & 78.78 \\
                & Conditional   & 64.15 & 71.80 & 70.65 & 65.26 & 80.32 & 70.43 \\
    \hline
        MMD     & Unconditional & 4,189 & 1,507 & 3,125 & 4,125 & 2,639 & 3,117 \\
                & Conditional   & 4,422 & 1,567 & 3,223 & 4,383 & 2,729 & 3,265 \\
    \hline
        ECD     & Unconditional & 2,390 & 6,043 & 369   & 366   & 349  \\
                & Conditional   & 2,394 & 8,057 & 1,270 & 413   & 649   &     &  \\
    \hline
  \end{tabular}
  \caption{Quantitative evaluation of our generative models at resolution 256 to the ground truth. For conditional generation we use the bounding boxes of the test set}
  \label{results-256}
\end{table*}

\end{appendices}
